# Efficient Informative Sensing using Multiple Robots

**Amarjeet Singh**                                         AMARJEET@EE.UCLA.EDU
**Andreas Krause**                                         KRAUSEA@CALTECH.EDU
**Carlos Guestrin**                                        GUESTRIN@CS.CMU.EDU
**William J. Kaiser**                                      KAISER@EE.UCLA.EDU

## Abstract

The need for efficient monitoring of spatio-temporal dynamics in large environmental applications, such as the water quality monitoring in rivers and lakes, motivates the use of robotic sensors in order to achieve sufficient spatial coverage. Typically, these robots have bounded resources, such as limited battery or limited amounts of time to obtain measurements. Thus, careful coordination of their paths is required in order to maximize the amount of *information* collected, while respecting the resource constraints. In this paper, we present an efficient approach for near-optimally solving the NP-hard optimization problem of planning such informative paths. In particular, we first develop *eSIP* (efficient Single-robot Informative Path planning), an approximation algorithm for optimizing the path of a single robot. Hereby, we use a Gaussian Process to model the underlying phenomenon, and use the mutual information between the visited locations and remainder of the space to quantify the amount of information collected. We prove that the mutual information collected using paths obtained by using *eSIP* is close to the information obtained by an optimal solution. We then provide a general technique, *sequential allocation*, which can be used to extend *any* single robot planning algorithm, such as *eSIP*, for the multi-robot problem. This procedure approximately generalizes any guarantees for the single-robot problem to the multi-robot case. We extensively evaluate the effectiveness of our approach on several experiments performed in-field for two important environmental sensing applications, lake and river monitoring, and simulation experiments performed using several real world sensor network data sets.

## 1. Introduction

Global climate change and corresponding impetus on sustainable practices for environment-related activities has brought forth the challenging task of observing natural phenomena exhibiting dynamics in both space and time. Observing and characterizing these dynamics with high fidelity will be critical for answering several questions related to policy issues for monitoring and control and understanding biological effects on activity of microbes and other organisms living in (or dependent on) these environments. Monitoring algal bloom growth in lakes and salt concentration in rivers, as illustrated in Fig. 1, are specific examples of related phenomena of interest to biologists and other environment scientists (MacIntyre, 1993; Ishikawa & Tanaka, 1993; MacIntyre, Romero, & Kling, 2002).

Monitoring environmental phenomena, such as algal bloom growth in a lake, requires measuring physical processes, such as nutrient concentration, wind effects and solar radiation, among others, across the entire spatial domain. One option to acquire data about such processes would be to statically deploy a set of sensing buoys (Reynolds-Fleming, Fleming, & Luettich, 2004). Due to the large spatial extent of the observed phenomena, this approach would require a large number of sensors in order to obtain high fidelity data. The spatio-temporal dynamics in these environments

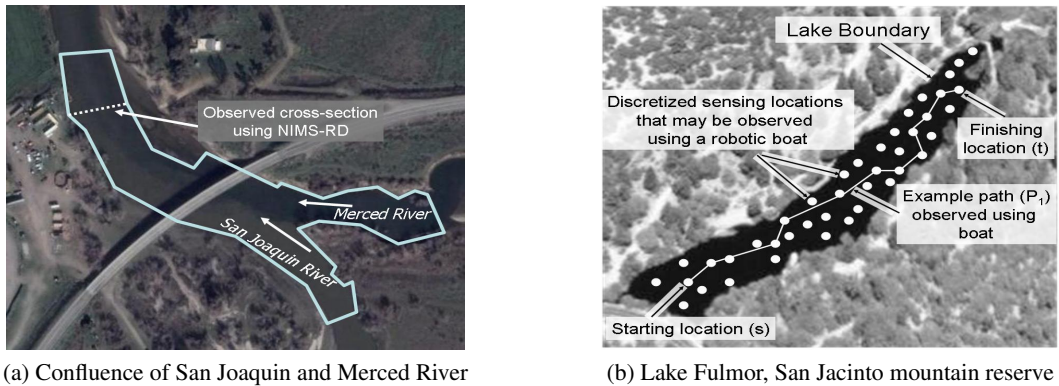

(a) Confluence of San Joaquin and Merced River    (b) Lake Fulmor, San Jacinto mountain reserve

Figure 1: Deployment sites used for performing path planning in-field.

motivate the use of actuated sensors – robots carrying sensors together with an efficient approach for planning the paths of these actuated sensors. These actuated sensors have been used in the past (Dhariwal et al., 2006) for measuring the phenomena at various locations and hence providing the biologists with critical information about the state of the lake.

Typically however, such robots have strict resource constraints, such as storage battery energy, that limits the distance they can travel or the number of measurements they can acquire before the observed phenomena varies significantly. These constraints necessitate careful motion planning for the robots – coordinating their paths in order to maximize the amount of collected information, while satisfying the given resource constraints. In this paper, we tackle this important problem of seeking *informative paths* for a collection of robots, subject to constraints on the cost incurred by each robot, e.g. due to limited battery capacity.

In order to optimize the paths of these robots, we first need to quantify the informativeness of any particular chosen path. In this work, we adopt an approach from spatial statistics and employ probabilistic models of the spatial phenomena. Using these models, informativeness can be viewed in terms of the uncertainty about our prediction of the phenomena at unobserved locations, given the observations made by the mobile robots at a subset of locations (the selected path). In particular, we use a rich class of probabilistic models called *Gaussian Processes* (GPs) (Rasmussen & Williams, 2006) that has been shown to accurately model many spatial phenomena (Cressie, 1991), and apply the *mutual information* (MI) criterion (Caselton & Zidek, 1984) to quantify the reduction in uncertainty achieved through selected robot paths.

Unfortunately, the problem of finding an optimal collection of paths, maximizing the mutual information criterion, is an **NP**-hard search problem, which is typically intractable even for small spatial phenomena. In this paper, we will develop an approximation algorithm which efficiently finds a provably near-optimal solution to this optimization problem. The key insight which will allow us to obtain such an algorithm is that the mutual information (and several other notions of informativeness (as discussed in Krause and Guestrin, 2007) satisfies *submodularity*, an intuitive diminishing returns property - making a new observation helps more if we have made only a few observations so far, and less if we have already made many observations (Krause et al., 2008).

The problem of optimizing the path of a *single robot* to maximize a submodular function over the visited locations was studied by Chekuri and Pal (2005), who developed an algorithm, *recursive-greedy*, with strong theoretical approximation guarantees. Unfortunately, the running time of their approach is *quasi-polynomial*: it scales as $M^{\log M}$, for $M$ possible sensing locations. This property makes the algorithm impractical for most environmental sensing applications, with typical numbers ($M$) of observation locations reaching to several hundreds and more. In this paper, we present two techniques – spatial decomposition and branch and bound search – for overcoming these limitations of the *recursive-greedy* approach of Chekuri et al., making it practical for real world sensing problems. We call this efficient approach for single robot path planning e*SIP* (efficient Single-robot Informative Path planning).

We then provide a general approach, *sequential-allocation*, which can be used to extend *any* single robot algorithm, such as e*SIP*, to the multi-robot setting. We furthermore prove that this generalization only leads to minimal reduction (independent of the number of mobile robots) of the approximation guarantee provided by the single robot algorithm. We combine e*SIP* with *sequential-allocation* to develop the first efficient path planning algorithm (e*MIP*) that coordinates multiple robots, each having a resource constraint, in order to obtain highly informative paths, i.e. paths that maximize any given submodular function, such as mutual information. By exploiting submodularity, we prove strong theoretical approximation guarantees for our algorithm.

We extensively evaluate the effectiveness of our approach on several experiments performed in-field for two important environmental sensing applications, lake and river monitoring. The river campaign was executed at the confluence of two rivers, Merced river and San Joaquin river, in California from August 7-11, 2007. Fig. 1a displays an aerial view of the San Joaquin deployment site. The lake campaign was executed at a lake located at the University of California, Merced campus from August 10-11, 2007. Fig. 1b displays an aerial view of lake Fulmor. In both campaigns, the *Networked Info Mechanical System (NIMS)* (Jordan et al., 2007), a cable based robotic system, was used to perform path planning while observing a two dimensional vertical plane (cross-section). In addition to analyzing data from these deployments, we provide extensive experimental analysis of our algorithm on several real world sensor network data sets, including the data collected using a robotic boat at lake Fulmor (Dhariwal et al., 2006).

This manuscript is organized as follows. We formally introduce the Multi-robot Informative Path Planning (MIPP) problem in Section 2. In Section 3, we discuss the *sequential-allocation* approach for extending any single robot path planning algorithm to the multi-robot setting while preserving approximation guarantees. We then review the *recursive-greedy* algorithm proposed by Chekuri et al. (Section 5), an example of such a single-robot algorithm. Subsequently, we present our spatial decomposition (Section 6) and branch and bound techniques (Section 7) which drastically improve the running time of *recursive-greedy* and make it practical for real world sensing applications. In Section 8, we evaluate our approach through in-field experiments as well as in simulations on real world sensing datasets. In Section 9, we review related work, and we present our conclusions in Section 10. The proofs for all our results are presented in the Appendix.

## 2. The Multi-robot Informative Path Planning Problem

We now formally define the *Multi-robot Informative Path Planning* (MIPP) problem. We assume that the spatial domain of the phenomenon is discretized into finitely many sensing locations $\mathcal{V}$. For each subset $\mathcal{A} \subseteq \mathcal{V}$, let $I(\mathcal{A})$ denote the *sensing quality*, i.e. the informativeness, of observing the phenomenon at locations $\mathcal{A}$. Details on appropriate choices for the sensing quality I are given below. We also associate with each location $v \in \mathcal{V}$, a *sensing cost* $C(v) > 0$, quantifying the expenses of obtaining a measurement at location $v$. When traveling between two locations, $u$ and $v$, a robot in-curs a *traveling cost* $C(u, v) > 0$. A robot traverses a *path* in this space: an $s$–$t$-path $\mathcal{P}$ is a sequence of $l$ locations starting at node $s$, and finishing at $t$. The cost $C(\mathcal{P})$ of path $\mathcal{P} = (s = v_1, v_2, \ldots, v_l = t)$ is the sum of sensing costs and traveling costs along the path, i.e. $C(\mathcal{P}) = \sum_{i=2}^{l-1} C(v_i) + \sum_{i=2}^{l} C(v_{i-1}, v_i)$. In the case $l = 2$, cost of the path $\mathcal{P}$ will only involve traveling cost between the starting and finishing locations $C(s, t)$. We will use the notation $\mathcal{P}$ to both refer to the sequence of nodes in the path, and to the subset of sensing locations $\mathcal{P} \subseteq \mathcal{V}$ (ignoring their sequence). For a collection of $k$ paths $\mathbf{P} = \{\mathcal{P}_1, \ldots, \mathcal{P}_k\}$, one for each robot, $\mathrm{I}(\mathbf{P}) = \mathrm{I}(\mathcal{P}_1 \cup \cdots \cup \mathcal{P}_k)$ denotes the sensing quality of the paths, which quantifies the amount of information collected by the $k$ paths. The goal of the MIPP problem is to find a collection $\mathbf{P}$ of $k$ paths, with specified starting and finishing location $s_i$ and $t_i$ (not necessarily different), such that each path has bounded cost $C(\mathcal{P}_i) \leq B$ for some specified budget $B$, and that the paths are the *most informative*, i.e. $\mathrm{I}(\mathbf{P})$ is as large as possible.

Formally, the problem can be defined as:

$$\max_{P_i \subseteq \mathcal{V}} \mathrm{I}(\cup_{i=1}^{k} \mathcal{P}_i); \quad \text{subject to } C(P_i) \leq B, \forall\, i \in \{1, \ldots, k\}. \tag{1}$$

In our lake monitoring example with the goal of performing surface monitoring using boats, we first discretized the two-dimensional surface of the lake into finitely many sensing locations (as depicted in Fig. 1b). For the single robot scenario, we then seek to find the most informative path $\mathcal{P}_1$ (in terms of predicting the algal bloom content) starting from location $s$ and finishing in location $t$. The experiment cost $C(v_i)$ corresponds to the energy required for making chlorophyll and related measurements (indicators of amount of algal bloom). The traveling cost $C(v_{i-1}, v_i)$ corresponds to the energy consumption when traveling from location $v_{i-1}$ to $v_i$. The budget $B$ quantifies the total energy stored in the boat's battery.

## 2.1 Quantifying Informativeness:

How can we quantify the sensing quality I? To model spatial phenomena, a common approach in spatial statistics is to use a rich class of probabilistic models called *Gaussian Processes* (GPs, *c.f.,* Rasmussen and Williams, 2006). Such models associate a random variable $\mathcal{X}_v$ with each location $v \in \mathcal{V}$. The joint distribution $P(\mathcal{X}_{\mathcal{V}})$ can then be used to quantify uncertainty in the prediction $P(\mathcal{X}_{\mathcal{V} \setminus \mathcal{A}} \mid \mathcal{X}_{\mathcal{A}} = \mathbf{x}_{\mathcal{A}})$ of phenomena at unobserved locations $\mathcal{X}_{\mathcal{V} \setminus \mathcal{A}}$, after making observations $\mathcal{X}_{\mathcal{A}} = \mathbf{x}_{\mathcal{A}}$ at a small subset $\mathcal{A}$ of locations. To quantify this uncertainty we use, for example, the *mutual information* (MI) criterion (as discussed by Caselton and Zidek, 1984). For a set of locations, $\mathcal{P}$, the MI criterion is defined as:

$$\mathrm{MI}(\mathcal{A}) \equiv H(\mathcal{X}_{\mathcal{V} \setminus \mathcal{A}}) - H(\mathcal{X}_{\mathcal{V} \setminus \mathcal{A}} \mid \mathcal{X}_{\mathcal{A}}) \tag{2}$$

where $H(\mathcal{X}_{\mathcal{V} \setminus \mathcal{A}})$ is the entropy of the unobserved locations $\mathcal{V} \setminus \mathcal{A}$, and $H(\mathcal{X}_{\mathcal{V} \setminus \mathcal{A}} \mid \mathcal{X}_{\mathcal{A}})$ is the conditional entropy of locations $\mathcal{V} \setminus \mathcal{A}$ after sensing at locations $\mathcal{A}$. Hence mutual information measures the reduction in uncertainty at the unobserved locations. Therefore, in our lake monitoring example, we would like to select the locations that most reduce the uncertainty in the algal bloom content prediction for the lake environment. Conveniently, in a GP, the mutual information criterion can be computed efficiently and analytically (Caselton & Zidek, 1984). The effectiveness of mutual information to select informative sensing locations was studied by Krause et al. (2008). Several alternative information criteria such as entropy (Ko et al., 1995), information disk model (Bai et al., 2006) and alphabetical optimality criterion such as A-, D- and E-optimal have also been used to associate sensing quality with observation locations in related problem domain.

## 2.2 Submodularity:

Even if we do not consider the constraints on the length of the paths of the robots, the problem of selecting locations that maximize mutual information is **NP**-hard (Krause et al., 2008). Hence, in general, we most likely cannot expect to be able to efficiently find the optimal set of locations. Instead, our goal will be to efficiently find *near-optimal* solutions, for which the sensing quality (e.g. mutual information), is provably close to the optimal sensing quality.

The key observation, which will allow us to obtain such strong approximation guarantees, is that mutual information satisfies the following diminishing returns property (Krause et al., 2008): The more locations we have already sensed, the less information we will gain by sensing a new location. This intuition is formalized by the concept of *submodularity*: A function $f$ is *submodular* (Nemhauser et al., 1978) if:

$$\forall \mathcal{A} \subseteq \mathcal{B} \subseteq \mathcal{V} \text{ and } s \in \mathcal{V} \setminus \mathcal{B}; \ \ f(\mathcal{A} \cup s) - f(\mathcal{A}) \geq f(\mathcal{B} \cup s) - f(\mathcal{B}). \tag{3}$$

Another intuitive property is that sensing quality is *monotonic*[1], which means that $\mathrm{I}(\mathcal{A}) \leq \mathrm{I}(\mathcal{B})$ for all $\mathcal{A} \subseteq \mathcal{B} \subseteq \mathcal{V}$. Hence, as we select more and more sensing locations, we will collect more and more information. Lastly, mutual information is normalized, i.e. $\mathrm{I}(\emptyset) = 0$.

We thus define our MIPP problem as the problem of optimizing paths of length at most $B$ for $k$ robots, such that the selected sensing locations maximize a *normalized, monotonic submodular function* $\mathrm{I}(\cdot)$. This definition of the MIPP problem allows our approach to be applied to *any* monotonic submodular objective function, not just mutual information. This generalization is very useful, as several other notions of informativeness can be shown to satisfy submodularity (Krause & Guestrin, 2007).

## 2.3 Online vs Offline Path Planning:

Many robotic path planning applications, such as search and rescue, involve uncertain environments with complex dynamics that can only be partially observed. Informative path planning – selecting the best locations to observe subject to given sensing constraints, in such uncertain environments necessitates a trade off between *exploration* (gathering information about the environment) and *exploitation* (using the current belief about the state of the environment most effectively). We distinguish two different classes of algorithms: nonadaptive (offline) algorithms, that plan and commit to the paths before any observations are made, and adaptive (online) algorithms, that update and replan as new information is collected. Both the online and offline settings are **NP**-hard optimization problems. In this paper, we only discuss the approximation algorithms for the offline setting that exploit the known belief about the environment for efficient path planning. We plan to work towards extending our approach for an *exploration-exploitation* trade-off to incorporate online model adaptation in the future.

## 3. Approximation Algorithm for MIPP

The problem of optimizing the path of a *single robot* (i.e. $k = 1$) to maximize a submodular function of the visited locations, constrained by an upper bound ($B$) on the path cost, was first studied by Chekuri and Pal (2005). We will review their *recursive-greedy* algorithm in detail in Section 5.

---

1. This monotonicity holds only approximately for mutual information (Krause et al., 2008), which however is sufficient for all purposes of this paper.

---

1 **Algorithm:** sequential-allocation

   **Input**: $B$, $k$, starting / finishing locations $s_1, \ldots, s_k, t_1, \ldots, t_k, \mathcal{V}$

   **Output**: A set of informative paths $\mathcal{P}_1, \ldots, \mathcal{P}_n$

2 **begin**

3    $\mathcal{A}_0 \leftarrow \emptyset$;

4    **for** $1 \leq i \leq k$ **do**

      `// Performing path planning for the` $i^{th}$ `robot`

5       $\mathcal{P}_i \leftarrow SPP(s_i, t_i, B, \mathcal{A}_{i-1}, \mathcal{V})$;

      `// Committing to the previously selected locations`

6       $\mathcal{A}_i \leftarrow \mathcal{A}_{i-1} \cup \mathcal{P}_i$;

7    **return** $\mathcal{P}_1, \ldots, \mathcal{P}_k$;

8 **end**

---

**Algorithm 1**: Sequential allocation algorithm for multi robot path planning using any single robot path planning algorithm SPP. Output set of paths $\mathcal{P}_1, \ldots, \mathcal{P}_k$ provides an approximation guarantee of $1 + \eta$ where $\eta$ is the approximation guarantee of single robot path planning algorithm $SPP$.

In our lake monitoring problem, we seek to plan *multiple* paths, one for each robot. One possibility is to apply the single-path algorithm to the *product graph*, i.e. plan a path over tuples of locations simultaneously representing the locations of *all* robots. However, such straightforward application of the single-robot planning algorithm would lead to an increase in running time which is exponential in the number of robots, and therefore intractable in practice. We are not aware of any sub-exponential approximation algorithm for this challenging multiple-robot path planning problem. In this paper, we present a simple algorithm for the multi-robot scenario that can exploit any approximation algorithm for the single robot case, such as the *recursive-greedy* algorithm, as discussed by Chekuri and Pal (2005), and (almost) preserve the approximation guarantee, while avoiding the exponential increase in running time.

Our algorithm, *sequential-allocation*, successively applies the single robot path planning algorithm $k$ times to get the paths for $k$ robots. Hereby, when planning the $j$th path, the approach takes into account the locations already selected by the previous $j - 1$ paths. Committing to the (approximately) best possible path at each stage before moving on to the next stage makes our approach "greedy" in terms of paths.

The pseudocode of the algorithm is presented in Algorithm 1 and Fig. 2 illustrates the approach for three robots. The algorithm takes as input the budget constraint $B$, number of available robots $k$, starting and finishing location for each available robot $s_1, \ldots, s_k, t_1, \ldots, t_k$ and the complete set of discrete observation locations $\mathcal{V}$ to select from. Let us assume that we have a single robot path planning algorithm, $SPP$, that takes as input a starting location $s_i$, a finishing location $t_i$, budget constraint $B$, a set of locations already selected for observation and a set of possible observation locations that can be visited. In Fig. 2, all the three robots have same starting and finishing location.

While planning the path for the first robot ($i = 1$), the input set of already selected observation locations is empty. At each subsequent stage, we commit to the locations selected in all the previous stages and pass the already observed locations as input to our next call to $SPP$. Let $\mathcal{A}_{i-1}$ be the locations already visited by paths $\mathcal{P}_1, \ldots, \mathcal{P}_{i-1}$, and $\mathcal{A}_0 = \emptyset$. Then the *residual information*, $\mathrm{I}_{\mathcal{A}_{i-1}}$ for a path $\mathcal{P}$ over unvisited locations is defined as $\mathrm{I}_{\mathcal{A}_{i-1}}(\mathcal{P}) = \mathrm{I}(\mathcal{A}_{i-1} \cup \mathcal{P}) - \mathrm{I}(\mathcal{A}_{i-1})$. It can be verified that if I is a normalized, monotonic and submodular function, then so is the residual information $I_{\mathcal{A}_{i-1}}$. Thus, at stage $i$ we use $SPP$ to find the most informative path with respect to the modified residual sensing quality function. In Fig. 2, when planning $\mathcal{P}_2$, locations selected for $\mathcal{P}_1$ are considered and the sensing quality function used is $I_{\mathcal{P}_1}$. Similarly, while evaluating the path $\mathcal{P}_3$, locations selected for $\mathcal{P}_1$ and $\mathcal{P}_2$ are taken into account and the sensing quality function used is $I_{\mathcal{P}_1 \cup \mathcal{P}_2}$.

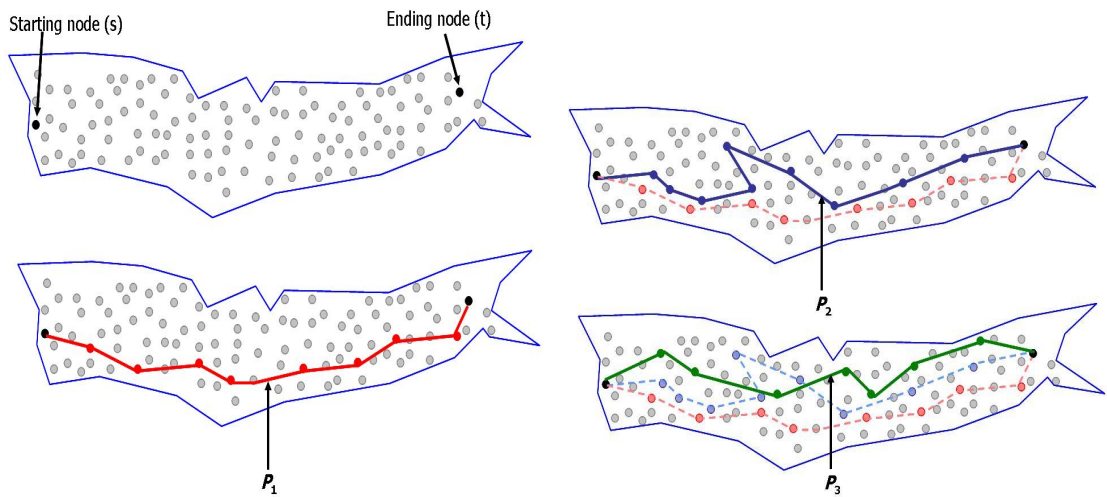

Figure 2: Illustration of sequential allocation algorithm for three robots, each with the same starting and finishing location.

Perhaps surprisingly, this straight-forward "greedy" sequential allocation approach is guaranteed to perform almost as well as the black box algorithm used for path planning. More formally, assume we have an $\eta$-approximate algorithm for the single robot problem, i.e. an algorithm which, starting with budget $B$ and a monotonic submodular function $f$, is guaranteed to find a path recovering at least a fraction of $1/\eta$ of the optimal information achievable with the same budget. In this case, the following theorem proves that the sequential allocation procedure has approximation guarantee close to $\eta$ as well:

**Theorem 1.** *Let $\eta$ be the approximation guarantee for the single path instance of the informative path planning problem. Then our sequential-allocation algorithm achieves an approximation guarantee of $(1 + \eta)$ for the MIPP problem. In the special case, where all robots have the same starting ($s_i = s_j, \forall i, j$) and finishing locations ($t_i = t_j, \forall i, j$), the approximation guarantee improves to $1/(1 - \exp(-1/\eta)) \leq 1 + \eta$.*

The work by Blum et al. (2003) proved Theorem 1 for the special case of additive (modular) sensing quality functions. In this paper, we extend their result to general submodular functions.

As an example of an $\eta$-approximate algorithm for the single robot problem, in the next section, we review the *recursive-greedy* algorithm as proposed by Chekuri and Pal (2005). This algorithm has an approximation guarantee $\eta$ of $\mathcal{O}(\log_2 |\mathcal{P}^*|)$, where $|\mathcal{P}^*|$ is the number of locations visited by an optimal solution $\mathcal{P}^*$. Hence, for this algorithm, the performance guarantee obtained for the MIPP problem through sequential allocation is $\mathcal{O}(\log_2 |\mathcal{P}^*|)$ as well[2].

---

2. In order to apply sequential allocation to the *recursive-greedy* algorithm, we can, when planning the $i$th path, simply pass the set of nodes visited by the previous $i - 1$ paths as the input parameter $\mathcal{R}$, as is illustrated in Algorithm 2.

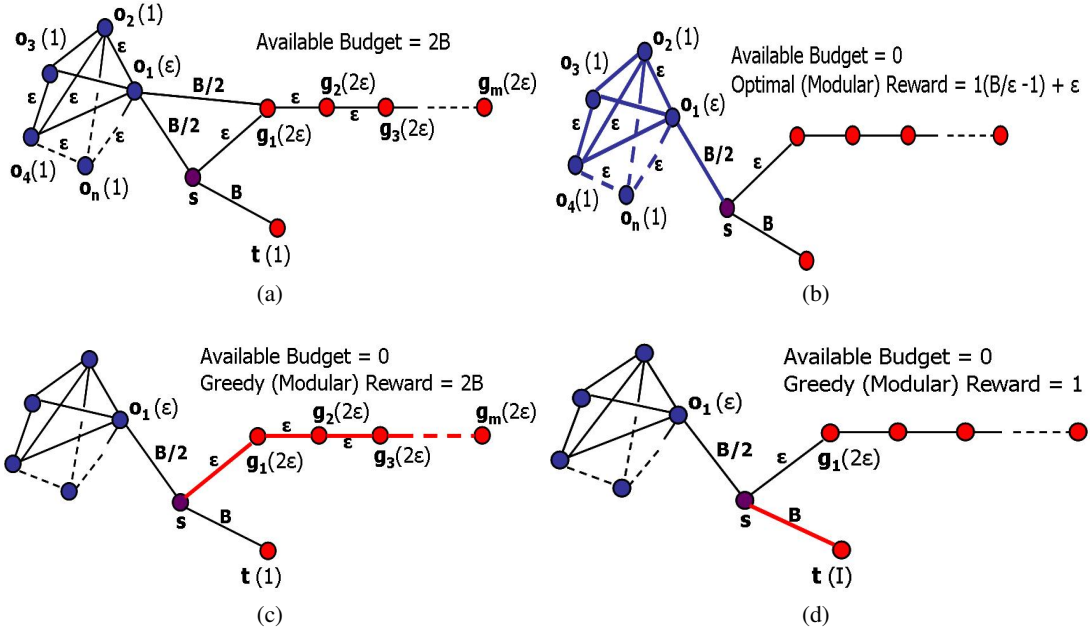

Figure 3: Illustration of performance of simple greedy approaches compared to an optimal approach.

## 4. A Note on Greedy Path Planning

The work by Krause et al. (2008) considered the sensor placement problem, where a subset $\mathcal{A} \subseteq \mathcal{V}$ of $k$ locations is selected in order to maximize the mutual information, without considering path costs. By exploiting the submodularity property of MI, they proved that if the discretization $\mathcal{V}$ is fine enough and the GP satisfies mild regularity conditions, greedily selecting locations based on this criterion is near-optimal. More specifically, the greedy algorithm (which we call *GreedySubset* in the following), after selecting the first $i$ locations $\mathcal{A}_i$, picks the location with maximum residual information i.e. $v_{i+1} = \operatorname{argmax}_v \mathrm{I}_{\mathcal{A}_i}(\{v\})$ and sets $\mathcal{A}_{i+1} = \mathcal{A}_i \cup \{v_{i+1}\}$. *GreedySubset* hence iteratively adds locations which increase mutual information the most. Using a result proposed by Nemhauser et al. (1978) on the performance of the greedy algorithm for submodular functions, the work by Krause et al. (2008) showed that *GreedySubset* selects sets which achieve mutual information of at least $(1 - 1/e)\operatorname{OPT} - \varepsilon$, where OPT is the optimal mutual information among all sets of the same size, and $\varepsilon$ is a small error incurred due to the discretization.

The strong performance of the greedy algorithm in the unconstrained (no traveling costs between locations) case motivates the question of whether a simple greedy approach could perform well in the more complex path planning setting considered in this paper. While it is difficult to give a general impossibility statement for such a question, several natural extensions of the greedy algorithm can be shown to perform arbitrarily badly.

For example, consider a setting where we define the cost $C(\mathcal{A})$ of a set of nodes as the cost of the cheapest path connecting the nodes $\mathcal{A}$. Assuming locations $\mathcal{A}_i$ have already been picked, a natural extension of the greedy algorithm will be to add a location $v$ which most improves the benefit-cost ratio

$$v^* = \operatorname*{argmax}_{v \in \mathcal{V} \setminus \mathcal{A}} \frac{\mathrm{I}_{\mathcal{A}_i}(v)}{C_{\mathcal{A}_i}(v)},$$

where $C_{\mathcal{A}_i}(v) = C(\mathcal{A}_i \cup \{v\}) - C(\mathcal{A}_i)$ is the increase in cost after adding $v$ to the already selected locations $\mathcal{A}_i$.

Fig. 3 shows a small example illustrating that this intuitive greedy procedure can perform arbitrarily poorly compared to an optimal approach. The example is illustrated in Fig. 3a, with $s$ as both the starting and the finishing location and $2B$ as the total available budget. The reward associated with each observation location is displayed in parenthesis with the corresponding locations. For the ease of illustration, we assume that the reward associated with each observation location is some modular function (instead of a submodular function). Traveling cost is associated with the corresponding edges in the example. Starting at location $s$, possible options for the first observation location are to select either of $o_1$, $g_1$ or $t$. Observation location $o_1$ will lead to a cluster of $n$ (= $B/\epsilon$) locations each separated by a traveling cost $\epsilon$ and with an associated reward of 1 (except $o_1$ that has an associated reward of $\epsilon$). $o_1$ is separated from $g_1$ by a traveling cost of $B/2$ while the rest of the locations in the cluster are assumed to be unreachable from any other location outside the cluster. Observation location $g_1$ will lead to a series of $m$ (= $B/\epsilon$) locations, each separated from the previous one by $\epsilon$ traveling cost and with an associated reward of $2\epsilon$.

As illustrated in Fig. 3b, an optimal approach would select $o_1$ as the first location, paying a traveling cost of $B/2$ and earning a very small reward $\epsilon$. Once the robot observes $o_1$, it can then observe the rest of $(B/\epsilon - 1)$ locations in the cluster, each providing a reward of 1 and return back to $s$ while spending a total of $2B$ as the traveling cost. Thus, the total reward collected by an optimal approach, for this example, will be $1(B/\epsilon - 1) + \epsilon$.

As illustrated in Fig. 3c, a "greedy" approach based on the reward-cost ratio will select $g_1$ as the first observation location (with the highest reward to cost ratio of 2). Since $o_1$ is at a distance $B/2$ away from $g_1$ and only provides a reward of $\epsilon$, this approach will continue along the series, observing all the locations till $g_m$ and returning back to $s$. Total reward collected by such an approach will be $2B$. On the other hand, a simple "greedy" approach based on reward (as illustrated in Fig. 3d) will simply select $t$ as the first observation location and return back to $s$, collecting a total reward of 1. Since the ratio $B/\epsilon$ can be arbitrarily large as $\epsilon \to 0$, the reward collected by simple intuitive greedy approaches ($2B$ or 1) can be arbitrarily poor when compared to the reward collected by an optimal approach ($1(B/\epsilon - 1) + \epsilon$).

Although, the reward function considered in the example was assumed to be a modular function, the submodular optimal reward can also be arbitrarily large, compared to submodular reward collected by simple greedy approaches (the difference between the submodular and modular reward will depend on the correlation of the selected observation locations). This insight necessitates the development of more complex algorithms for path planning as considered in this paper.

## 5. The Recursive-greedy Algorithm

We will now review the *recursive-greedy* algorithm as proposed by Chekuri and Pal, since it forms the basis for our efficient single robot path planning approach. The basic strategy of the algorithm is a divide-and-conquer approach. Any path from the starting location ($s$) to finishing location ($t$) has a middle location ($v_m$) such that there are same number of locations (or different by at most 1) on either side of $v_m$ in the $s - t$ path. Thus, the problem of finding a $s - t$ path can be divided into two smaller subproblems of finding smaller subpaths ($s - v_m$ and $v_m - t$) and then concatenating these small subpaths. While having the same *number* of locations, the subpaths on either side of the middle node can have different *costs*, i.e. the budget for the total path has to be split into two smaller

---

1    **Algorithm:** recursive-greedy ($RG$)

     **Input**: $s,t,B,\mathcal{R},iter$

     **Output**: An informative path $\mathcal{P}$

2    **begin**

3      **if** $c(s,t) > B$ **then**

4         **return** Infeasible;

5      $\mathcal{P} \leftarrow s,t$;

6      Base case: $iter$=0 **return** $\mathcal{P}$;

7      $m \leftarrow f_{\mathcal{R}}(\mathcal{P})$;

     `// Trying each location as middle node`

8      **foreach** $v_m \in \mathcal{V}$ **do**

        `// Trying all possible budget splits`

9         **for** $1 \leq B_1 \leq B$ **do**

           `// Planning subpath on one side of the middle node`

10            $\mathcal{P}_1 \leftarrow RG(s,v_m,B_1,\mathcal{R},iter-1)$;

           `// Planning subpath on other side of the middle node,`
             `committing to nodes selected in first subpath`

11            $\mathcal{P}_2 \leftarrow RG(v_m,t,B-B_1,\mathcal{R} \cup \mathcal{P}_1,iter-1)$;

12            **if** $f_{\mathcal{R}}(\mathcal{P}_1 \cup \mathcal{P}_2) > m$ **then**

13               $\mathcal{P} \leftarrow \mathcal{P}_1 \cup \mathcal{P}_2$;

14               $m \leftarrow f_{\mathcal{R}}(\mathcal{P})$;

15      **return** $\mathcal{P}$;

16    **end**

---

**Algorithm 2**: Recursive greedy algorithm for single robot instance of MIPP as proposed by Chekuri and Pal (2005). Output path $\mathcal{P}$ provides an approximation guarantee of $I_X(\mathcal{P}) \geq I_X(\mathcal{P}^*)/\lceil 1 + \log k \rceil$, where I represent the submodular reward function, $\mathcal{P}^*$ represent an optimal path and $k$ represent the number of nodes in the optimal path.

budgets (not necessarily equal), one for each subpath. Searching for the best middle location and trying all possible budget splits on either side of the middle location, while optimizing the complete $s - t$ path, would result in an exhaustive search for the optimal solution and therefore will be prohibitively expensive. Instead of performing this exhaustive search, the *recursive-greedy* algorithm follows a simple greedy strategy, wherein for each of the possible budget splits and each possible middle nodes considered, one can *first* plan the optimal subpath on one side of the middle location, then commit to the planned subpath and optimize for the subpath on the other side. Such a path, consisting of independently optimized subpath $s - v_m$ and a subpath $v_m - t$ optimized subject to observation locations already selected in $s - v_m$, may result in a suboptimal $s - t$ path. Nonetheless, Chekuri and Pal proved that such a path has an approximation guarantee of $\mathcal{O}(\log_2 |\mathcal{P}^*|)$, where $|\mathcal{P}^*|$ is the number of locations visited by an optimal solution $\mathcal{P}^*$.

In order to implement such a greedy approach, the recursive calls planning the second subpath will – similarly as done in sequential allocation – optimize a residual reward function which measures the incremental gain taking into account the information already obtained by the locations selected in the first subpath. More formally, let the set $\mathcal{P}_1$ refer to the locations selected in the first subpath, and consider the *residual* submodular function $f_{\mathcal{P}_1}$ over a set of locations $\mathcal{A}$ as

$f_{\mathcal{P}_1}(\mathcal{A}) = f(\mathcal{A} \cup \mathcal{P}_1) - f(\mathcal{P}_1)$. If $\mathcal{P}_2$ is the set of locations in the second subpath, then it holds that $f(\mathcal{P}_1) + f_{\mathcal{P}_1}(\mathcal{P}_2) = f(\mathcal{P}_1 \cup \mathcal{P}_2)$. Hence, if the first recursive call (with submodular function $f$) returns path $\mathcal{P}_1$, and the second recursive call (with submodular function $f_{\mathcal{P}_1}$) returns path $\mathcal{P}_2$, then the sum of the scores of the subproblems exactly equals the score of the concatenated path.

Let us now formalize the intuitive description of the *recursive-greedy* algorithm. The pseudocode of the algorithm is presented in Algorithm 2. The inputs to the algorithm are a starting location $s$, a finishing location $t$, an upper bound on the path cost $B$, a parameter $\mathcal{R}$ that defines the residual for the submodular function such that the function that needs to be maximized is defined as $f_{\mathcal{R}}(\mathcal{P}) = f(\mathcal{P} \cup \mathcal{R}) - f(\mathcal{R})$, and a parameter $i$ that represents the recursion depth. The maximum number of locations that can be selected at each stage is calculated using the recursion depth as $2^i$. In the base case (recursion depth $i = 0$), the algorithm simply returns the path $\mathcal{P} = (s, t)$ (if the cost $c(s, t) \leq B$).

In the recursive case, the algorithm searches for a $s - t$ path with maximum reward by iterating over all possible locations (that can be reached with given budget constraint) as middle locations (Line 8), i.e. locations that could possibly split the required path into two subpaths with equal number of locations on either side. For each such middle location, the algorithm explores all possible splits of available budget (Line 9) across the two subpaths on either side of the middle location. Reducing the recursion depth by 1, for each subpath, ensures that the same number of locations are selected on either side of the middle location. However, before exploring the second subpath, the algorithm commits to the locations selected in the first subpath by passing them as input through the "residual" parameter (Line 11). The two subpaths found in such a way are then concatenated to provide a complete $s - t$ path. The algorithm stores the best possible $s - t$ path over the already searched problem space, replacing it with a better path whenever such a path is found.

## 5.1 Structure of the Search Problem

It is instructive to consider the recursive structure generated by the *recursive-greedy* algorithm. Fig. 4 illustrates an example of such a structure when running *recursive-greedy* for our lake sensing application with the given starting ($s$) and finishing ($t$) location and an upper bound on the path cost ($B$). The search using *recursive-greedy* can be represented graphically as a *sum-max* tree. At the root is a *max* node representing the objective of finding a $s - t$ path with maximum possible reward, while the cost of the path is bounded by budget $B$. For each such max node, the children in the search tree represent *sum* nodes corresponding to sum of rewards collected from the two subpaths on either side of the middle location. Therefore, at the end of the first iteration, the graphical representation will have a *max* node as root with several *sum* nodes as children, for each feasible middle location and each possible budget splits around that middle location. A partial tree at the end of first iteration is shown in Fig. 4a.

For each *sum* node, formed at the end of the first iteration, the algorithm is then applied recursively on the left subpath. Thus the first step of second iteration seeks to find a $s - v_m$ path with maximum possible reward under the budget constraint corresponding to the respective budget split for the *sum* node. Then, their approach commits to the selected locations on the left side, and recurses on the right subpath (to search for a $v_m - t$ path), given these selected locations. As a result, each *sum* node will have two *max* nodes as children, each representing an objective to find a subpath of maximum reward on either side of the selected middle location. This algorithm is "greedy" in that it commits to the locations selected in the first subpath before optimizing the second subpath.

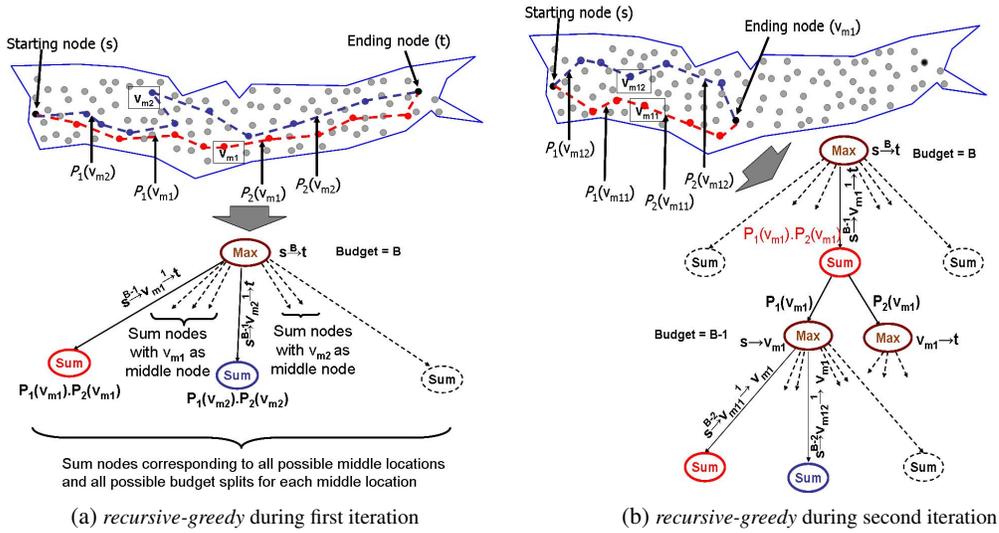

(a) *recursive-greedy* during first iteration      (b) *recursive-greedy* during second iteration

**Figure 4:** Illustration of recursive greedy algorithm, as proposed by Chekuri and Pal, for the lake sensing application. Sum-max tree presents the graphical representation of the problem space.

A partial tree at the end of second iteration is shown in Fig. 4b. Despite the greedy nature, the *recursive-greedy* approach provides the following approximation guarantee:

**Theorem 2.** *(Chekuri & Pal, 2005)* Let $\mathcal{P}^* = (s = v_0, v_1, \ldots, v_k = t)$ *be an optimal s-t-path solution. Let* $\mathcal{P}$ *be the path returned by* $RG(s, t, B, \mathcal{R}, i)$. *If* $i \geq \lceil 1 + \log k \rceil$, *then* $I_X(\mathcal{P}) \geq I_X(\mathcal{P}^*)/\lceil 1 + \log k \rceil$.

Hence, the *recursive-greedy* solution $\mathcal{P}$ obtains at least a fraction of $\frac{1}{\lceil 1 + \log_2 k \rceil}$ of the optimal information, where $k \leq n$, i.e. the total number of locations traversed by the optimal path will be smaller than the total number of locations in the discretized spatial domain. Referring back to Theorem 1, for the MIPP problem using *recursive-greedy* as the single robot path planning approach, $\eta = \lceil 1 + \log k \rceil$.

## 5.2 Running Time

By inspecting the recursive structure, the running time of the *recursive-greedy* algorithm can be seen to be *quasi-polynomial*. More specifically, the running time of the algorithm is $\mathcal{O}((MB)^{O(\log_2 M)})$, where $B$ is the budget constraint and $M = |\mathcal{V}|$ is the total number of possible observation locations. So, even for a small problem with $M = 64$ locations, the exponent will be 6, resulting in a very large computation time, making the algorithm impractical for observing several real world physical processes.

The large computational effort required by *recursive-greedy* can be attributed to two issues: 1) the large branching factor at each of the *max* nodes of the recursion tree (*sum* nodes for each possible middle node and each possible budget split across that middle node) and 2) (possibly) unnecessary recursion while exploring subtrees in problem space that can not provide us with an improved reward compared to current best solution. In the following sections, we propose two complementary approaches (can be used independently of the others) which are intended to ameliorate these concerns: a spatial decomposition technique, and a branch and bound approach. Spatial decomposition (discussed in Section 6) seeks to reduce the high branching factor (i.e. the number of *sum* nodes in the search tree) by clustering the sensing locations and then running the *recursive-greedy* over these clusters instead of actual sensing locations. Branch and bound (discussed in Section 7) seeks to avoid unnecessary recursion by maintaining a lower and an upper bound on the possible reward from a subtree in the search tree and pruning the tree accordingly. These two approaches, together with *sequential-allocation* (discussed in Section 3) provide an efficient algorithm for multi robot informative path planning.

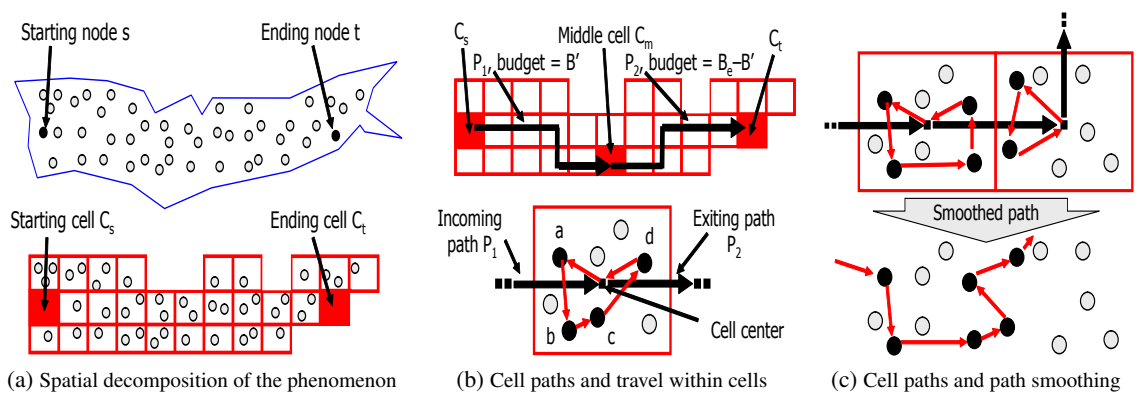

Figure 5: Illustration of spatial decomposition in *recursive*-e*SIP* using surface sensing in lake environment as an example. The sensing domain ((a), top) is decomposed into a grid of cells ((a), bottom). *recursive*-e*SIP* jointly optimizes over cell-paths ((b), top) and allocations of experiments in the cells ((b), bottom). Within the cells, locations are connected to cell center. *recursive*-e*SIP* concatenates paths between-cell and within cell paths ((c), top) and finally heuristics are applied in e*MIP* to smooth the path ((c), bottom).

## 6. Spatial Decomposition – Approximating MIPP as SD-MIPP

In this section, we explain in detail the process of spatial decomposition and the corresponding improvements in running time achieved through this process. Our approach assumes that the traveling cost between arbitrary locations is given by their euclidean distance.

An intuitive approach for improving the running time is to spatially decompose the sensing region into smaller sub-regions, each containing a cluster of sensing locations. We can thus think about planning informative paths as deciding which sub-regions to explore, and then deciding which locations to sense within these sub-regions. The idea of exploring the sub-regions motivates the decomposition of the sensing domain into smaller regions (cells). We can then run the *recursive-greedy* algorithm on these cells instead of the actual sensing locations. Since the size of each cellular region is small, traveling cost within each cell can be ignored[3]. Once we ignore the traveling cost within the cells, sensing locations inside the selected cells can be chosen using the *GreedySubset* approach (as proposed by Krause et al., 2008), taking advantage of its strong approximation guar-antee in an unconstrained setting as discussed in Section 4. Fig. 5 presents an illustration of our approach and is explained as follows:

1. We decompose the sensing region, containing finitely many discrete sensing locations (*c.f.,* Fig. 5a, top), into a collection of non-overlapping cells $\widetilde{\mathcal{V}} = \{\mathcal{C}_1, \mathcal{C}_2, \ldots, \mathcal{C}_N\}$ (*c.f.,* Fig. 5a, bottom). The distance between two cells is defined as the distance between the centroids of these cells. Each cell $\mathcal{C}_i$ contains a set of locations $v_i \in \mathcal{V}$, representing sensing locations, such that the coordinates of these locations, in a euclidean metric space, lie within the boundary of the containing cell.

2. We approximate the original MIPP problem with the *spatially decomposed MIPP problem, or SD-MIPP problem* on $\widetilde{\mathcal{V}}$. In SD-MIPP, we *jointly* optimize over cell-paths in $\widetilde{\mathcal{V}}$ (*c.f.,* Fig. 5b, top) using the *recursive-greedy* algorithm, and over the *allocation* of observations within the cells visited by the paths using the *GreedySubset* algorithm. Thus, when allocating measurements to a cell, we ignore the traveling cost within the cell (*c.f.,* Fig. 5b, bottom). Since the cells are not very large, this simplification only leads to a small additional cost when the SD-MIPP solution is transformed back to the original MIPP problem.

3. We transfer the (approximate) SD-MIPP solution, consisting of a cell-path and an allocation of observations to cells (*c.f.,* Fig. 5c, top), back to the original MIPP problem. We then smooth the path (*c.f.,* Fig. 5c, bottom) using heuristics, e.g. the *tour-opt* heuristics as discussed by Lin (1965).

---

3. There may be robotic platforms where non-holonomic motion constraints will make small motions much more challenging and thus traveling cost for smaller distances within a cell may become non-negligible. For such systems, with large traveling cost for smaller motions, some system specific constraints may be possible to account for while performing cellular decomposition or the greedy algorithm may be constrained to not select locations that are "too" close).

Dual optimization of cell paths and budget allocation for observations within each visited cell motivated splitting the available budget $\widetilde{\mathcal{B}}$ into a budget $B_t$ for traveling between the cells and a budget $B_e$ for making experiments at sensing locations within the visited cells. Such a split can be easily incorporated in *recursive-greedy* algorithm as well but was not required as the paths in *recursive-greedy* were optimized over observation locations and not cells containing these locations. Formally, the SD-MIPP problem is the following: We want to find a path $\mathcal{P}_C^* = (\mathcal{C}_s = \mathcal{C}_{i_1}, \ldots, \mathcal{C}_{i_l} = \mathcal{C}_t)$, for each robot $i$ with starting cell $\mathcal{C}_s$ containing the starting node $s$ and finishing cell $\mathcal{C}_t$ containing the finishing node $t$, with a travel cost of at most $B_t$. This travel budget is measured in terms of distances between centers of visited cells, and the cost of traveling *within* the cells is defined as 0. In addition, for each visited cell $\mathcal{C}_{i_j}$ in $\mathcal{P}_C^*$, we want to select a set of sensing locations $\mathcal{A}_{i_j}$, such that the total experimental cost (for making observations within the visited cells) is upper bounded by $B_e$, i.e. $C(\mathcal{A}_{i_1} \cup \cdots \cup \mathcal{A}_{i_l}) \leq B_e$, and that the information $\mathrm{I}(\mathcal{A}_{i_1} \cup \cdots \cup \mathcal{A}_{i_l})$ is as large as possible. The optimal SD-MIPP solution uses the optimal split of the budget $\widetilde{\mathcal{B}}$ into $B_t$ and $B_e$. To simplify the presentation, we rescale the costs such that the cells form a uniform grid of quadratic cells with width $L$, and assume that the sensing cost $C_{exp}$ is constant over all locations. These assumptions can easily be relaxed, but they allow us to relate the path costs to the number of cells traversed, to simplify the discussion.

The following lemma states that there exists an SD-MIPP version ($\mathcal{P}_C^*$) of the MIPP-optimal path ($\mathcal{P}^*$), with (almost) the same cost, and the same information.

**Lemma 3.** *Let* $\mathcal{P}^* = (s = v_0, v_1, \ldots, v_l = t)$ *be an optimal s-t-path solution to MIPP, constrained by budget $B$. Then there exists a corresponding SD-MIPP path* $\mathcal{P}_C^* = (\mathcal{C}_s = \mathcal{C}_{i_1}, \ldots, \mathcal{C}_{i_l} = \mathcal{C}_t)$, *traversing through locations* $\mathcal{A}_{i_1} \cup \cdots \cup \mathcal{A}_{i_l}$, *with budget $\widetilde{\mathcal{B}}$ of at most $2\sqrt{2}B + 4L$ collecting the same information.*

---

1    **Algorithm:** e*MIP*

     **Input**: $\widetilde{\mathcal{B}}$, $k$, starting / finishing locations $s_1, \ldots, s_k, t_1, \ldots, t_k$

     **Output**: A collection of informative paths $\mathcal{P}_1, \ldots, \mathcal{P}_k$

2    **begin**

3        Perform spatial decomposition into cells;

4        Find starting and ending cells $\mathcal{C}_{s_i}$ and $\mathcal{C}_{t_i}$;

5        $\mathcal{R} \leftarrow \emptyset$;

       `// Path planning for each robot`

6        **for** $i = 1$ **to** $k$ **do**

           `// Trying different combination of traveling and`
              `experimental budget`

7           **for** $iter = 0$ **to** $\lfloor \log_2 \widetilde{\mathcal{B}} \rfloor$ **do**

8              $B_e \leftarrow \widetilde{\mathcal{B}} - 2^{iter}$;

9              $\mathcal{P}'_{iter} \leftarrow$ *recursive*-e*SIP* ($\mathcal{C}_{s_i}, \mathcal{C}_{t_i}, B_e, \mathcal{R}, iter$);

10             Smooth $\mathcal{P}'_{iter}$ using *tour-opt* heuristics;

11           $\mathcal{P}_i \leftarrow \operatorname{argmax}_{iter} \mathrm{I}(\mathcal{P}'_{iter})$;

12           $\mathcal{R} \leftarrow \mathcal{R} \cup \mathcal{P}_i$;

13        **return** $\mathcal{P}_1, \ldots, \mathcal{P}_k$;

14    **end**

---

**Algorithm 3**: e*MIP* algorithm for informative multi robot path planning. Procedure from Line 7 to Line 11 effectively implements e*SIP* algorithm. e*SIP* is then repeated (Line 6) using *sequential allocation* described in Section 3 (Line 6) to get paths for each robot $i$.

We now present an algorithm for finding an approximately optimal solution to SD-MIPP, and then we show that this solution gives us an approximate solution to the original MIPP problem, with just slightly increased cost of $2\sqrt{2}B + 4L$, for ensuring that the optimal solution for MIPP exists in the corresponding SD-MIPP setting.

## 6.1 Algorithm for SD-MIPP

Our e*MIP* algorithm solves the SD-MIPP problem on $\widetilde{\mathcal{V}}$ and then smooths out the paths over the selected observation locations to provide a solution to MIPP. Let us first clarify the algorithmic nomenclature specifically:

- *recursive*-e*SIP*: implements an approach similar to *recursive-greedy* for selecting a path over $\widetilde{\mathcal{V}}$ and greedily selects the observation locations within each visited cell using *GreedySubset*;

- e*SIP*: iterates through different values of traveling budget by calling *recursive*-e*SIP* with corresponding values of input $B_e$ and $i$ and smoothing the output path from *recursive*-e*SIP* using *tour-opt* heuristics;

- e*MIP*: effectively implements *sequential-allocation* with e*SIP* as the single robot path planning algorithm

The complete algorithm works as follows: An outer loop (Line 6 in Algorithm 3) implements the sequential allocation algorithm for performing path planning for multiple robots. The procedure inside the outer loop (Line 7 to Line 11 in Algorithm 3) implements the e*SIP* algorithm. This procedure iterates through different combination of traveling and experimental budget, allocating $B_t$ ($= 2^{iter}$) out of the total budget $\widetilde{\mathcal{B}}$ for traveling between the cells, and $B_e (= \widetilde{\mathcal{B}} - B_t)$ for making experiments within the visited cells. Stepping through $B_t$ in powers of 2 results in faster performance ($\log_2 \widetilde{\mathcal{B}}$ instead of $\widetilde{\mathcal{B}}$ iterations). If we increase the input budget $\widetilde{\mathcal{B}}$ by a factor of 2, the exponential increase in traveling budget is guaranteed to try traveling budget, $B_t$ ($= 2^{iter} \geq B_{tApp}$) where $B_{tApp}$ is the traveling budget for the best approximation path. Since the overall budget $\widetilde{\mathcal{B}}$ is increased by a factor of 2, the remaining experimental budget is also guaranteed to be more than the experimental budget corresponding to the best approximation path. Therefore, exponential increase in traveling budget will only increase the required budget $\widetilde{\mathcal{B}}$ by at most a factor of 2. The e*SIP* procedure then calls *recursive*-e*SIP* (explained in Algorithm 4), selecting the cells to visit, and greedily allocating observations in the visited cells. Finally, the e*SIP* procedure calls *tour-opt* heuristics to smooth the output path from *recursive*-e*SIP*.

The *recursive*-e*SIP* procedure takes as input a starting cell $\mathcal{C}_s$, a finishing cell $\mathcal{C}_t$, an experimental budget $B_e$, a residual $\mathcal{R}$ indicating the locations visited thus far (initially passed empty from e*MIP*), and a maximum recursion depth, $iter$ (initially passed $\log_2 B_t$ from e*MIP*). We then:

1. Iterate through all possible choices of middle cells $\mathcal{C}_m$ (such that there are, almost, equal number of cells on either side of $\mathcal{C}_m$) and budget splits $\widetilde{B_e}$ (of the available experimental budget $B_e$) to spend for making experiments on the subpaths from $\mathcal{C}_s$ to $\mathcal{C}_m$ and $\mathcal{C}_m$ to $\mathcal{C}_t$ (*c.f.,* Fig. 5b). The budget splits $\widetilde{B_e}$ can either be linearly (more accurate) or exponentially (faster) spaced, as described below.

2. Recursively find a subpath $\mathcal{P}_1$ from $\mathcal{C}_s$ to $\mathcal{C}_m$, constrained by budget $B'$, leaving the remaining budget $(B_e - B')$ for the other subpath $\mathcal{P}_2$. Reducing recursion depth ($iter$) by 1, for each of the subpaths $\mathcal{P}_1$ and $\mathcal{P}_2$, ensures that equal number of cells are visited on either side $\mathcal{C}_m$. The lowest level of recursion depth 0 signifies the cell selected for the corresponding path. At the lowest recursion level, we then use the *GreedySubset* algorithm (*c.f.,* Section 4) to select the sensing locations based on the residual information function $\mathrm{I}_{\mathcal{R}}$ and constrained by budget $B'$. As an illustration, the black locations in the middle cell $\mathcal{C}_m$ in Fig. 5b bottom, are selected by the *GreedySubset* algorithm with budget $B' = 4$ such that they provide the maximum improvement in mutual information.

3. We then commit to the locations selected in $\mathcal{P}_1$, and recursively find a subpath $\mathcal{P}_2$ from $\mathcal{C}_m$ to $\mathcal{C}_t$, with experimental budget $B_e - B'$. Committing to the locations selected in $\mathcal{P}_1$ requires that we greedily select the sensing locations at lowest recursion level based on the residual information function $\mathrm{I}_{\mathcal{R} \cup \mathcal{P}_1}$.

4. Finally, we concatenate the locations obtained in $\mathcal{P}_1$ and $\mathcal{P}_2$ to output the best path from the algorithm (*c.f.,* Fig. 5c, top).

## 6.2 Linear vs. Exponential Budget Splits

Step 1 of the *recursive*-e*SIP* procedure (as explained in Section 6.1) considers different budget splits $B' \in \widetilde{B_e}$ to the left and right subpaths. Similar to the recursive greedy algorithm, one can choose $\widetilde{B_e} = \{0, 1, 2, 3, \dots, B_e - 1, B_e\}$ to be linearly spaced. Since the branching factor is proportional to the number of considered splits, linear budget splits leads to a large amount of computation effort.

---

1 **Algorithm:** *recursive-eSIP*

  **Input**: $\mathcal{C}_s, \mathcal{C}_t, B_e, \mathcal{R}, iter$

  **Output**: An informative path $\mathcal{P}$ from $\mathcal{C}_s$ to $\mathcal{C}_t$

2 **begin**

3   **if** $(d(C_s, C_t) > 2^{iter} L)$ **then return** *Infeasible*;

    // Greedy node selection within starting and finishing cell

4   $\mathcal{P} \leftarrow GreedySubset_{B_e, \mathcal{R}}(v_i : v_i \in \mathcal{C}_s \cup \mathcal{C}_t)$;

5   **if** *(iter = 0)* **then return** $\mathcal{P}$;

6   $reward \leftarrow I_{\mathcal{R}}(\mathcal{P})$;

    // Trying each cell as middle cell

7   **foreach** $\mathcal{C}_m \in \mathcal{C}$ **do**

      // Trying each possible budget split

8     **for** $B' \in \widetilde{\mathcal{B}_e}$ **do**

        // Planning subpath on one side of the middle cell

9       $\mathcal{P}_1 \leftarrow$ *recursive-eSIP* $(\mathcal{C}_s, \mathcal{C}_m, B', \mathcal{R}, iter - 1)$;

        // Planning subpath on other side of the middle cell

          while committing to nodes selected in first subpath

10       $\mathcal{P}_2 \leftarrow$ *recursive-eSIP* $(\mathcal{C}_m, \mathcal{C}_t, B_e - B', \mathcal{R} \cup \mathcal{P}_1, iter - 1)$;

11       **if** $(I_{\mathcal{R}}(\mathcal{P}_1.\mathcal{P}_2) > reward)$ **then**

12         $\mathcal{P} \leftarrow \mathcal{P}_1.\mathcal{P}_2$;

13         $reward \leftarrow I_{\mathcal{R}}(\mathcal{P})$;

14   **return** $\mathcal{P}$;

15 **end**

**Algorithm 4**: *recursive-eSIP* procedure for path planning.

An alternative is to consider only exponential splits: $\widetilde{\mathcal{B}_e} = \{0, 2^0, 2^1, 2^2, \ldots, 2^{\log_2 B_e}\} \cup \{B_e, B_e - 2^0, B_e - 2^1, B_e - 2^2, \ldots, 0\}$. In this case, the branching factor is only logarithmic in the experimental budget. Even though we are not guaranteed to find the same solutions as with linear budget splits, we can both theoretically (as given by Lemmas 4 and 7) and empirically (as illustrated in Fig. 14c and 14d) show that the performance only gets slightly worse in this case, compared to a significant improvement in running time. In addition to these two ways of splitting the budget, we also considered one-sided exponential budget splits (i.e. $\widetilde{\mathcal{B}_e} = \{0, 2^0, 2^1, 2^2, \ldots, 2^{\log_2 B_e}\}$), which further reduces the branching factor by a factor of 2 compared to the exponential splits defined above. Although we do not provide theoretical guarantees for this third possibility, we experimentally found it to perform very well (*c.f.,* Section 8).

### 6.3 Algorithmic Guarantees

Our algorithm is greedy in two ways:

- At recursion depth 0, the sensing locations are selected greedily based on the mutual information criterion.

- Before exploring the subpath $\mathcal{P}_2$, *recursive-eSIP* procedure commits to the locations selected in subpath $\mathcal{P}_1$.

Due to the these greedy steps, *recursive*-e*SIP* is an approximation algorithm and does not necessarily find an optimal solution. The following lemma, however, guarantees a performance bound for the path output by the e*SIP* procedure:

**Lemma 4.** *Let* $\mathcal{P}_C^* = (\mathcal{C}_s = \mathcal{C}_1, \dots, \mathcal{C}_k = \mathcal{C}_t)$ *be an optimal solution for single robot instance of SD-MIPP, constrained by budget* $\widetilde{\mathcal{B}}$, *where an optimal set of locations are selected within each visited cell* $\mathcal{C}_j$. *Let* $\widehat{\mathcal{P}}$ *be the solution returned for e*SIP. *Then* $\mathrm{I}(\widehat{\mathcal{P}}) \geq \frac{1-1/e}{1+\log_2 k} \mathrm{I}(\mathcal{P}_C^*)$.

## 6.4 Solving the MIPP Problem

Now, we need to transfer the approximately optimal solution obtained for SD-MIPP back to MIPP. A path over cells, with observation locations selected greedily within each visited cell, is transformed into a path over observation locations by connecting all locations selected in cell $\mathcal{C}_{i_j}$ to the cell's center, (as indicated in Fig. 5b bottom), then connecting all selected centers to a path (Fig. 5c top), and finally expanding the resulting tree into a tour by traversing the tree twice (by traversing each edge of the tree once in each direction, a set of nodes connected by a tree can be converted into a set of nodes connected by a path). This traversal results in a tour which is at most twice as long as the shortest tour connecting the selected vertices. (Of course, an even better solution can be obtained by applying an improved approximation algorithm for TSP, such as the algorithm proposed by Christofides, 1976). The following Theorem completes the analysis of our algorithm:

**Theorem 5.** *Let* $\mathcal{P}^*$ *be the optimal solution for the single robot instance of the MIPP problem with budget constraint B. Then, our e*SIP *algorithm will find a solution* $\widehat{\mathcal{P}}$ *achieving an information value of at least* $\mathrm{I}(\widehat{\mathcal{P}}) \geq \frac{1-1/e}{1+\log_2 N} \mathrm{I}(\mathcal{P}^*)$, *whose cost is no more than* $2(2\sqrt{2}B + 4L)(1 + L\frac{\sqrt{2}}{C_{exp}})$ *in the case of linear budget split for* $\widetilde{\mathcal{B}_e}$ *and no more than* $2(2\sqrt{2}B + 4L)(1 + L\frac{\sqrt{2}}{C_{exp}})N^{\log_2 \frac{3}{2}}$ *in the case of exponential budget split for* $\widetilde{\mathcal{B}_e}$.

The performance guarantee is w.r.t. the number of *cells* $N$ instead of the number $M$ of sensing locations, as was the case in the work by Chekuri and Pal (2005). However, the input budget constraint is violated by an amount based on the size of cells during the spatial decomposition. This violation in input budget constraint leads to a tradeoff between computation effort and additional cost incurred that can be tuned based on specific application requirements. If the size of the cell is small (in the limit reducing each cell to each observation location), the number of cells will be large and will result in higher computation time with reduced additional cost. On the other hand, if the size of the cell is large, the computation time will be small and the algorithm needs to pay higher additional traveling cost.

Running time analysis of e*SIP* is straightforward. The algorithm calls the routine *recursive*-e*SIP* $\log_2 B$ times. If $T_\mathrm{I}$ is the time to evaluate the mutual information I, then the time for computing greedy subset $T_{gs}$ (Line *4*, Algorithm 4) is $\mathcal{O}(N_C^2 \ T_\mathrm{I})$, where $N_C$ is the maximum number of locations per cell. At each recursion step we try all the cells that can be reached with the available traveling budget (Line *7*, Algorithm 4). For the possible experimental budget split, we try all (linearly or exponentially spaced) splits of $B_e \in \widetilde{\mathcal{B}_e}$ among the two subpaths $\mathcal{P}_1$ and $\mathcal{P}_2$ (Line *8*, Algorithm 4). The recursion depth would be $\log_2(\min(N, \widetilde{\mathcal{B}}))$. The following proposition states the running time for e*SIP*:

**Proposition 6.** *The worst case running time of e*SIP *for linearly spaced splits of the experimental budget is* $\mathcal{O}\left(T_{gs}\log_2 B(NB)^{\log_2 N}\right)$*, while for the exponentially spaced splits of the experimental budget it is* $\mathcal{O}\left(T_{gs}\log_2 B(2N\log_2 B)^{\log_2 N}\right)$

Comparing this running time to the *recursive-greedy* algorithm ($\mathcal{O}((MB)^{O(\log_2 M)})$), we note a reduction from $B$ to $\log_2 B$ in the base, and log of the number of locations ($\log_2 M$) to log of the number of cells ($\log_2 N$) in the exponent. These two improvements turn the impractical *recursive-greedy* approach into a much more viable algorithm.

Varying the number of cells (and correspondingly the size of each cell) results in a trade-off between the computation effort and the traveling cost within the cell that is ignored by the e*SIP* algorithm. Proposition 6 states that the computation effort is directly proportional to the number of cells "N". Therefore as we increase the number of cells, corresponding computation effort for the e*SIP* algorithm will also increase. On the other hand, reducing the number of cells will result in increasing the size of each of the cell. Since e*SIP* algorithm ignores the traveling cost within the cell, larger cell size will imply larger traveling cost ignored by the e*SIP* algorithm and hence larger overshoot in the cost of the resultant output path over the input budget B. Lemma 3 states the corresponding additional cost incurred by the output path calculated using e*SIP* algorithm in terms of the cell size "L". Based on the specific application requirements, one can decide the appropriate number of cells and fine tune the trade-off between computation effort and additional path cost incurred. Fig. 14f shows that the corresponding collected reward did not vary significantly as we varied the number of cells for the application of observing temperature in a lake environment.

## 7. Branch and Bound

The spatial decomposition technique effectively enables a trade-off between running time complexity and achieved approximation guarantee. However, the e*SIP* algorithm still has to solve a super-polynomial, albeit sub-exponential, search problem. In the following, we describe several branch and bound techniques which allow further reduction in the computation effort making our approach tractable for real world sensing experiments.

### 7.1 Problem Representation

The specific structure of the search space representation motivated many of the proposed branch and bound approaches. Similarly to the recursive structure of the *recursive-greedy* algorithm (discussed in Section 5), the *recursive-*e*SIP* problem structure can also be represented as a *sum-max* tree, as shown in Fig. 6a. A small difference exists in the selection of observation locations along the solution path. In the case of *recursive-greedy*, each of the *sum* nodes traversed in the selected path represents a physical observation location. However, in the case of *recursive-*e*SIP*, each *sum* node in the selected path represents a cell in the corresponding traversed path. The observation locations at the *sum* node are selected greedily, within the corresponding cell, based on the available experimental budget. Using the *sum-max* tree problem structure, we now explain the proposed branch and bound approaches to prune parts of the tree that will not provide any further improvement over the currently known best solution path. All of the proposed branch and bound techniques are outlined in the *recursive-*e*SIP* procedure presented in Algorithm 5.

---

1    **Algorithm:** *recursive-*e*SIP* with branch and bound

    **Input:** $\mathcal{C}_s, \mathcal{C}_t, B_e, \mathcal{R}, iter, rewardLB, \alpha$

    **Output:** An informative path $\mathcal{P}$ from $\mathcal{C}_s$ to $\mathcal{C}_t$

2    **begin**

3      **if** $(d(\mathcal{C}_s, \mathcal{C}_t) > 2^{iter}L)$ **then**

4        **return** *Infeasible*

5      $\mathcal{P} \leftarrow GreedySubset_{B_e, \mathcal{R}}(v_i : v_i \in \mathcal{C}_s \cup \mathcal{C}_t)$;

6      **if** *(iter = 0)* **then**

7        **return** $\mathcal{P}$

8      $filterCells \leftarrow \cup \mathcal{C}_i \;\; \forall \mathcal{C}_i \; s.t. \; d(\mathcal{C}_s, \mathcal{C}_i) \leq 2^{iter}L/2 \;$ and $\; d(\mathcal{C}_i, \mathcal{C}_t) \leq 2^{iter}L/2$ ;

9      **foreach** $\mathcal{C}_m \in filterCells$ **do**

10        **for** $B' \in \widetilde{\mathcal{B}_e}$ **do**

           `// Calculating upper bound using GreedySubset`

11          $UB_{\mathcal{P}_1} \leftarrow calculateUB(\mathcal{C}_s, \mathcal{C}_m, B', iter - 1, \mathcal{R})$;

12          $UB_{\mathcal{P}_2} \leftarrow calculateUB(\mathcal{C}_s, \mathcal{C}_m, B_e - B', iter - 1, \mathcal{R})$;

13          **if** $((UB_{\mathcal{P}_1} + UB_{\mathcal{P}_2}) > \alpha * rewardLB)$ **then**

            `// Calculating lower bound for` $P_1$

14            $heur_{\mathcal{P}_1} \leftarrow$ heuristicOP$(\mathcal{C}_s, \mathcal{C}_m, B', \mathcal{R}, iter - 1)$;

15            $LB_{\mathcal{P}_1} \leftarrow max(\mathrm{I}_{\mathcal{R}}(heur_{\mathcal{P}_1}), rewardLB - UB_{\mathcal{P}_2})$;

            `// Recursive search for` $P_1$

16            $\mathcal{P}_1 \leftarrow$ *recursive-*e*SIP* $(\mathcal{C}_s, \mathcal{C}_m, B', \mathcal{R}, iter - 1, LB_{\mathcal{P}_1}, \alpha)$;

            `// Calculating lower bound for` $P_2$

17            $heur_{\mathcal{P}_2} \leftarrow$ heuristicOP$(\mathcal{C}_m, \mathcal{C}_t, B_e - B', \mathcal{R} \cup \mathcal{P}_1, iter - 1)$;

18            $LB_{\mathcal{P}_2} \leftarrow max(\mathrm{I}_{\mathcal{R} \cup \mathcal{P}_1}(heur_{\mathcal{P}_2}), rewardLB - \mathrm{I}_{\mathcal{R}}(\mathcal{P}_1))$;

            `// Recursive search for` $P_2$

19            $\mathcal{P}_2 \leftarrow$ *recursive-*e*SIP* $(\mathcal{C}_m, \mathcal{C}_t, B_e - B', \mathcal{R} \cup \mathcal{P}_1, iter - 1, LB_{\mathcal{P}_2}, \alpha)$;

20            **if** $(\mathrm{I}_{resid}(\mathcal{P}_1.\mathcal{P}_2) > rewardLB)$ **then**

21              $\mathcal{P} \leftarrow \mathcal{P}_1.\mathcal{P}_2$;

22              $rewardLB \leftarrow \mathrm{I}_{resid}(\mathcal{P}_1.\mathcal{P}_2)$;

23      **return** $\mathcal{P}$;

24 **end**

---

**Algorithm 5**: *recursive-*e*SIP* procedure with branch and bound approaches for efficient path planning. Each procedure corresponds to a *max* node in the search space with input $rewardLB$ representing the calculated lower bound. A *sum* node in the search space effectively combines the recursive calls to each of the subpaths (implemented in Line *16* and Line *19*). Since recursion reduces the traveling budget ($2^{iter}L$) by half, the initial pruning in Line *8* removes the cells that can not be reached in the next recursion step. Line *15* and Line *18* calculate the lower bound for subpaths on either side of the selected middle cell. Input $\alpha$ represents the scaling factor for one of the sub-approximation heuristics. Approximation guarantee for the output path $\mathcal{P}$ is given as $\mathrm{I}(\widehat{\mathcal{P}}) \geq \frac{1 - 1/e}{1 + \log_2 N} \mathrm{I}(\mathcal{P}^*)$ where I is the submodular reward function and $\mathcal{P}^*$ is the optimal path.

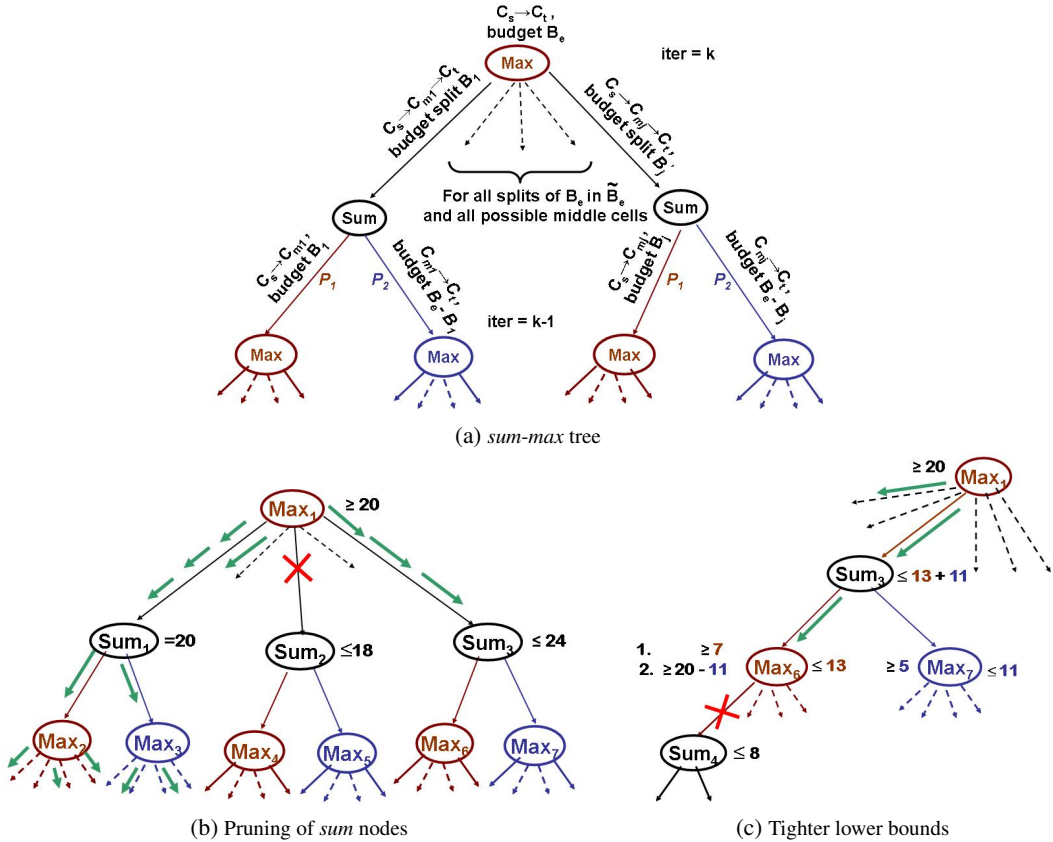

(a) *sum-max* tree

(b) Pruning of *sum* nodes

(c) Tighter lower bounds

**Figure 6:** Illustration of our branch & bound approach. (a) shows the *sum-max* tree representing the search space. Each *max* node selects a middle cell and a budget allocation, and each *sum* node combines two subpaths on either side of the selected middle cell. (b) shows how upper bound at a *sum* node (e.g. a value of 18 at $Sum_2$), when smaller than the lower bound of the parent *max* node (e.g. a value of 20 for $Max_1$) can be used to prune branches in the search tree. (c) shows how lower bound at a *max* nodes is tightened (e.g. a value of 7 at $Max_6$ is improved to 9 using upper bound of 11 at sibling $Maxn_7$ and lower bound of 20 at grandparent $Max_1$) to allow further pruning which otherwise may not have been possible (e.g. pruning of $Sum_4$ with upper bound value of 8).

## 7.2 Efficient Search of the Problem Space

In a naive implementation of *recursive-eSIP*, the entire recursion tree would eventually be traversed. However, many of the considered subpaths may be highly suboptimal. Several heuristics have been proposed in the past for similar path planning problem with empirical efficiency claims, but without any approximation guarantee. We use one such heuristic (*c.f.,* Chao et al., 1996, hereafter referred to as *heuristicOP*) to calculate a solution path satisfying the budget constraints, while trying to maximize the collected reward. Since such a path can be efficiently calculated with small computation effort, we use this path as an initial known solution. The total reward collected in this path is used as an input lower bound (input variable $rewardLB$ in Algorithm 5) for the root *max* node. Since the computation effort associated with *heuristicOP* is small, it is also used at the rest of the *max* nodes in the search tree to calculate the lower bound for these nodes (discussed in detail in Section 7.2.2).

For each of the child *sum* nodes, an upper bound for the collected reward is calculated by exploiting the submodularity of the reward function (procedure $calculateUB$ called in Line *11* and 12

---

1 **Algorithm:**calculateUB

   **Input**: $\mathcal{C}_s, \mathcal{C}_t, B_e, iter, \mathcal{R}$

   **Output**: An upper bound UB on information gain

2 **begin**

      // Selecting set of reachable cells

3      possibleCells $\leftarrow \cup \mathcal{C}_i \;\; \forall \mathcal{C}_i \; s.t. \; d(\mathcal{C}_s, \mathcal{C}_i) + d(\mathcal{C}_i, \mathcal{C}_t) \leq 2^{iter} L$ ;

      // Greedy node selection within reachable cells

4      $\mathcal{P} \leftarrow GreedySubset_{B_e, \mathcal{R}}(v_i : v_i \in possibleCells)$;

5      UB $\leftarrow \mathrm{I}_{resid}(\mathcal{P})$;

6      **return** $UB$;

7 **end**

---

**Algorithm 6**: Procedure for calculating upper bound at *max* nodes. Upper bound of child *max* nodes is added to obtain upper bound at parent *sum* node.

in Algorithm 5 and explained in detail in Algorithm 6). We then only need to process the *sum* node children with upper bounds greater than the current best solution (Line *13* in Algorithm 5). The current best solution for the parent *max* node is updated when the collected reward from any of the child *sum* nodes is greater than the previously known best solution reward (Line *20* in Algorithm 5).

Fig. 6b presents a graphical illustration of this concept. After completely exploring branch $\mathrm{Sum}_1$, the current best solution of value 20 is updated as a lower bound for $\mathrm{Max}_1$. A smaller lower bound (18) at $\mathrm{Sum}_2$ results in pruning of sub-branch rooted at $\mathrm{Sum}_2$. However, nodes such as $\mathrm{Sum}_3$ with upper bound (24) higher than the current best solution (20), need to be explored further as they can potentially provide a solution path with a better reward than the current best solution.

### 7.2.1 UPPER BOUND ON THE *Sum* NODES

Algorithm 6 presents the *calculateUB* procedure for obtaining an upper bound on the collected reward at each *max* node and is used in *recursive-eSIP* (Line *11*, 12 in Algorithm 5) for pruning the search space. The upper bound at a *sum* node is calculated by adding the upper bound of each of the child *max* nodes. We calculate the upper bounds by relaxing the path constraints, and then finding an optimal set of reachable locations for each path ($\mathcal{P}_1$ and $\mathcal{P}_2$). Since this problem itself is **NP**-hard, we exploit the submodularity of reward function and approximate it using the *GreedySubset* algorithm. Fig. 7 illustrates an example of calculating the upper bound. We first calculate the set of reachable locations w.r.t. the remaining traveling budget. These locations are contained within the cells $C_i$ reachable from cells $C_s$ and $C_t$ (Line *3* in Algorithm 6). Such a boundary for reachable locations is illustrated by an ellipse in Fig. 7.

Then, we run the *GreedySubset* algorithm to greedily select best possible $B_e$ locations from all the possible reachable locations (Line *4* of Algorithm 6). As an example, $\mathcal{V}_i$ and $\mathcal{V}_j$ are selected using *GreedySubset* in Fig. 7. Since *GreedySubset* guarantees a constant factor $(1 - 1/e)$ approximation (Nemhauser et al., 1978), multiplying the resulting information value by $(1 - 1/e)^{-1}$ provides an upper bound on the information achievable by the path (and hence the corresponding *max* child node). Therefore, in Fig. 7 the reward collected from locations $\mathcal{V}_i$ ($\mathrm{MI}(\mathcal{V}_i)$) and $\mathcal{V}_j$ ($\mathrm{MI}(\mathcal{V}_j)$) when multiplied by the factor $(1 - 1/e)^{-1}$ provides upper bound for the collected reward. However, since the path cost constraint are relaxed, the total cost of observing $\mathcal{V}_i$ and $\mathcal{V}_j$ ($d_{si} + d_{ij} + d_{jt}$) may be more than the available budget $B$. In Fig. 6c, for example, we use *calculateUB* to get upper bounds 13 for $\text{Max}_6$ and 11 for $\text{Max}_7$, resulting in an upper bound of $13 + 11 = 24$ for $\text{Sum}_3$[4].

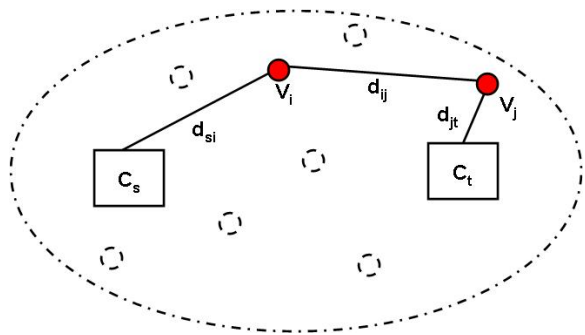

Figure 7: Illustration of calculating upper bound using *GreedySubset*.

### 7.2.2 LOWER BOUND ON THE *Max* NODES:

Effective pruning of subtree rooted at *sum* nodes would require calculating the lower bounds for the parent *max* node efficiently. One way to calculate such lower bounds is by exploring one branch completely (as explained in Section 7.2). This procedure will be computationally expensive. Instead, we implement two other ways for acquiring such lower bounds faster: Using *heuristicOP* [5] (as explained above for obtaining the initial best solution), and based on the current best solution of the grandparent *max* node. We then use the larger of two different lower bounds.

Fig. 6c illustrates the graphical presentation of the procedure for calculating the lower bounds using the current best solution of the grandparent *max* node. We call this procedure *altLB*. We calculate an upper bound (exploiting the submodularity) of 11 for $\text{Max}_7$ node. For the node $\text{Max}_6$, since the grandparent node $\text{Max}_1$ has a lower bound of 20, the subtree rooted at $\text{Max}_6$ has to provide a reward of at least 9 (20 - 11) to be explored further. The lower bound of value 9 calculated using *altLB* is tighter than the lower bound provided by the heuristic (7), and enabled pruning of branch $\text{Sum}_4$ (with upper bound 8).

Lines 15 and 18 in Algorithm 5 illustrate the *altLB* procedure. While using *altLB*, the lower bound for subpath $\mathcal{P}_1$ (in *Line 15*), is calculated using the upper bound of subpath $\mathcal{P}_2$. On the other hand, while calculating the lower bound using *altLB* for subpath $\mathcal{P}_2$ (in *Line 18*), the exact reward from $\mathcal{P}_1$ ($I_\mathcal{R}(\mathcal{P}_1)$) is used instead of the upper bound. Since the actual reward is always tighter than the calculated upper bound, the lower bound calculated for subpath $\mathcal{P}_2$ (using *altLB*) will be tighter than the lower bound calculated for subpath $\mathcal{P}_1$. This motivates exploring the subpath with higher experimental budget first such that the upper bound for the unexplored subpath (with lower experimental budget) is tighter making the lower bound for the first subpath tighter[6]. The heuristic for exploring the subpath with higher experimental budget first was also exploited to further improve the computation effort.

---

4. We can even compute tighter online bounds for maximizing monotonic submodular functions, as discussed by Nemhauser et al. (1978).

5. *heuristicOP* was only proposed for modular functions but we found it to provide good solution paths even in the submodular setting.

6. We note that with higher experimental budget, *GreedySubset* (used to calculate the upper bound) can potentially select more locations that are far apart (since the path cost constraint are ignored). When path cost constraint is incorporated, such locations will become infeasible and will make the upper bound loose.

Maintaining a lower bound at each node in the search tree also makes our approach *anytime*, i.e. the search can be terminated at any point even before it is completed. The current best solution from the graph already searched will be available after this early termination. Early termination is particularly advantageous in scenarios when it is required to obtain the best possible path traversed by the robot with a hard upper bound on the available time to calculate such a path.

### 7.2.3 NODE ORDERING

The illustration in Fig. 6b demonstrates that a better currently known solution will likely help in increased pruning of the search tree. In order to improve the current best solution faster, at each *max* node we explore the *sum* nodes in the decreasing order of their upper bounds. The intuitive idea is that a higher upper bound is a likely indicator for higher reward value. Thus the upper bound in Line *11* and 12 in Algorithm 5 can be calculated separately and the rest of the computation (in loops implemented in Line *9* and 10 in Algorithm 5) can then be executed in decreasing order of upper bound. Such an approach is similar to *node ordering* that is employed to improve the pruning efficiency of Depth First Branch and Bound (DFBnB) (Zhang & Korf, 1995).

### 7.2.4 SUB-APPROXIMATION

Upper and lower bounds derived as explained above can potentially be loose. We can address this issue, and further trade off collected information with improved execution time, by introducing several sub-approximation heuristics. As a first heuristic, once the *node ordering* is performed, we explore only the top $K$ *sum* nodes. This heuristic, termed as sub-approximation  (Ibaraki et al., 1983), is found to be effective in practice.

As a second heuristic, instead of comparing the lower bound of a parent *max* node directly with the upper bound from the child *sum* nodes (when deciding which subproblems to prune), we scale up the lower bound by a factor of $\alpha > 1$ (Line *13* of Algorithm 5). This scaling often allows us to prune many branches that would not have been pruned otherwise. Unfortunately, this optimistic pruning can also potentially cause us to prune branches that should not have been pruned, and decrease the information collected by the algorithm. In practice, for sufficiently small $\alpha$ values, this procedure can speed up the algorithm significantly, without much effect on the quality of the solution. This performance comparison for both computation effort and collected reward using several real world sensing datasets is discussed in Section 8.2.

## 8. Experimental Results

We performed several experiments both in-field as well as in simulation (using real world sensing datasets) to demonstrate the usefulness of our proposed algorithm for several diverse environmental sensing applications. In-field experiments were performed using the Networked InfoMechanical System (NIMS) (Jordan et al., 2007), a tethered robotic system. Real world sensing datasets used for performing scaling and multi robot experiments in simulation were collected using either a network of static sensors or a robotic boat.

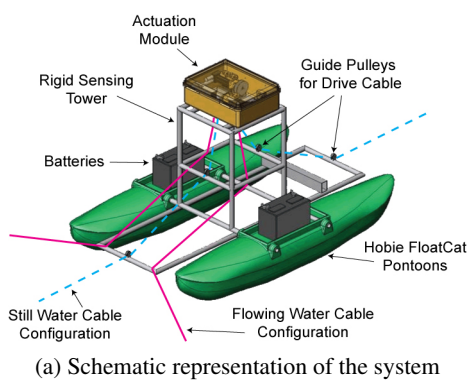

(a) Schematic representation of the system

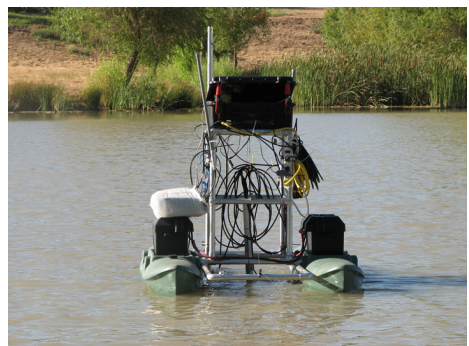

(b) Image captured while performing path planning

**Figure 8:** Aquatic based NIMS (NIMS-AQ)is a platform in the NIMS family used for performing path planning in the lake environment.

## 8.1 In-field Experiments

Several experiments were performed in-field to demonstrate the applicability of modeling a phenomenon as a *Gaussian Process* and using e*MIP* to perform path planning for diverse aquatic sensing applications. These include a river monitoring application with the objective of studying salt concentration, and lake monitoring for several applications of interest to limnologists. In each of these applications, NIMS was used to monitor a cross-section (two dimensional vertical plane in an environment) in the aquatic environment. The phenomenon of interest is then modeled as a *Gaussian Process* and we use the mutual information criterion as submodular reward function, quantifying the informativeness of observation locations. The learned Gaussian Process model and mutual information objective are then provided as input to e*MIP* and the subset of locations as output by the algorithm are subsequently observed, again using NIMS as the robotic platform. In order to quantify the efficiency of our approach, we predict the phenomenon at unobserved locations and compute the root mean square (RMS) error between the predicted phenomenon and the ground truth (calculated by observing at all the uniformly spaced locations before and after the path planning experiment).

### 8.1.1 ROBOTIC PLATFORM:

The Aquatic Networked InfoMechanical Systems platform (NIMS-AQ) is the latest in the family of NIMS systems (Jordan et al., 2007; Pon et al., 2005; Borgstrom et al., 2006), developed specifically for aquatic applications and used during the lake deployment. The family of NIMS systems had been successfully deployed for several terrestrial and aquatic sensing applications. In 2006 alone, NIMS was used in several successful campaigns in forests (La Selva, Costa Rica and James Reserve, California), rivers (San Joaquin, California and Medea Creek, California), lake (Lake Fulmor, California), and mountain ecosystems (White Mountains, California),

Fig. 8a displays the schematic view of the system. The basic infrastructure of the system includes a rigid sensing tower supported by two Hobie FloatCat pontoons[7] in a catamaran configuration. An actuation module resides on top of the sensing tower that drives the horizontal cable and vertical payload cable (horizontal and vertical motion respectively) across a cross-section of the aquatic environment. Power for the system is provided by two deep cycle marine batteries housed on top of the pontoons. The horizontal drive cable is kept center-aligned to the craft by using guide

---

7. Developed by Hobie Cat Company.

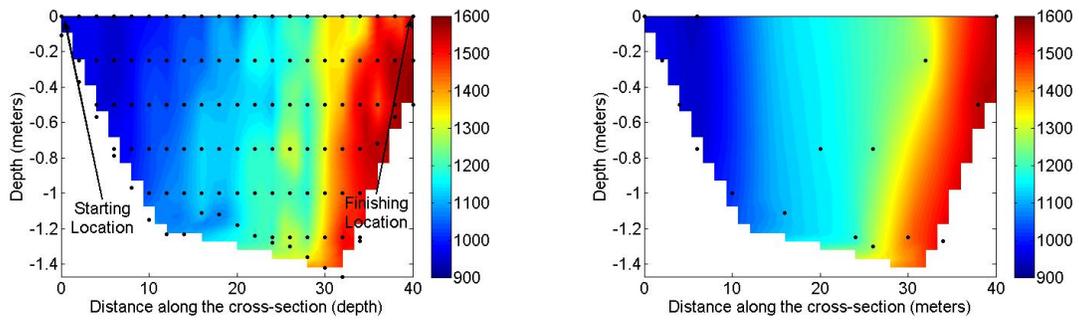

(a) Observed distribution during a raster scan on August 11

(b) Predicted distribution after observing at locations as output by e*MIP*

Figure 9: Distribution of electrical conductivity (microSiemens per centimeter) as observed at the confluence of San Joaquin river, California. Points represent observation locations during the corresponding experiment.

pulleys that can be repositioned based on the type of aquatic environment in which NIMS-AQ is sampling (flowing or still water conditions). Fig. 8b shows NIMS-AQ performing path planning in the lake environment.

### 8.1.2 SENSING IN A RIVER ENVIRONMENT

The first in-field application of our approach was executed at the confluence of two distinct rivers, Merced river and San Joaquin river, in California from August 7-11, 2007 (hereafter referred to as San Joaquin deployment). Fig. 1a displays an aerial view of the San Joaquin deployment site. The scientific objective at the confluence zone is to characterize the transport and mixing phenomena at the confluence of two distinct rivers – Merced river (relatively low salinity) and the agricultural drainage-impacted San Joaquin River (relatively high salinity) by observing several parameters that may indicate the mixing behavior of the two streams. Such river observations are useful for answering important questions pertaining to the spatio-temporal variability of velocity and water quality dynamics resulting from pollutant inputs, hydrodynamic mixing regimes, and biogeochemical cycling processes that are themselves distributed in time and space. Understanding such mixing patterns are important for policy issues related to water distribution from river ecosystems (Brekke et al., 2004).

The total width of the observed cross-section was 40 meters with the maximum depth of 1.4 meters (closer to the middle of the cross-section). Several experiments had been executed in the past to characterize the mixing phenomena at this confluence site (Singh et al., 2007a; Harmon et al., 2007). Primary experimental design during these campaigns comprised of making observations at uniformly spaced locations in a two dimensional cross-section (hereafter referred to as raster scan) and repeating these experiments several times to understand the spatial and temporal dynamics in the environment. Each of these experiments took several hours, thus restricting the experiments to a very small number of cross-sections (one or two) within the limited deployment time. However, a detailed understanding of the confluence environment would require observing multiple cross-sections, within the limited time frame. This necessitates the use of an adaptive sampling approach that can model the observed phenomenon, make observations at a small number of locations based on that model and then effectively predict the phenomenon at the unobserved locations.

Mixing patterns were characterized at the confluence by observing electrical conductivity that indicated the amount of salt concentration in the water. Fig. 9a displays typical distribution at a cross-section in the confluence zone with x-axis representing the distance along the cross-section and y-axis representing the depth. Low concentration of electrical conductivity towards the lower x values is contributed by clear water from the Merced river with the other end displaying high concentration of salts carried by the San Joaquin river. We first use the data from one such raster scan performed on the first day of the deployment (displaying similar characteristics) to learn a non-stationary Gaussian Process model, using a covariance function parameterization as described by Krause and Guestrin (2007). The parameters are chosen by maximizing the marginal likelihood (Rasmussen & Williams, 2006). This non-stationary process was learned by dividing the complete region into smaller sub-regions and combining the locally-stationary GPs from each of these sub regions.

A total of 114 locations were observed during the raster scan and used for learning the GP model. A set of 16 locations was selected out of the total of 114 (14%) using the e*MIP* algorithm with the starting and finishing location on either end of the cross-section as displayed in Fig. 9a. This set of 16 observation locations was then observed over the next few days. With the required dwelling time[8] of 30 seconds for observing electrical conductivity, large reduction in number of observation locations resulted in a significant reduction in experimental time as well (14% compared to the raster scan).

Since the environmental phenomena exhibit spatial and temporal dynamics, we performed raster scans before and after our experiment to get a measure of ground truth for electrical conductivity. The predicted electrical conductivity, as computed after making the observations at the subset of 16 locations selected using e*MIP*, is then compared with this ground truth. Fig. 9b displays the predicted distribution of specific conductivity with points representing the observed locations as output by e*MIP*. Fig. 9a displays the distribution as observed using raster scan performed just before the path planning experiment.

The RMS error between the predicted distribution and the raster scan performed before the path planning experiment was 45.99 $\mu S/cm$. On the other hand, the RMS error between the predicted distribution and the raster scan performed after the path planning experiment was 53.87 $\mu S/cm$. The RMS error between the two raster scans performed before and after the path planning experiment, indicating the temporal variation in the environment, was 57.55 $\mu S/cm$. Low RMS error for our predicted distribution, when compared with the RMS error between the raster scans performed before and after the path planning experiment clearly indicates the effectiveness of our approach for modeling and path planning in such environments. Path planning experiments performed during other days also demonstrated similar prediction accuracy, while maintaining the significant reduction in total experimental time.

### 8.1.3 SENSING IN A LAKE ENVIRONMENT

The second set of in-field experiments was executed at a lake on the campus of University of California, Merced from August 10-11, 2007 (hereafter referred to as lake deployment). This site was chosen based on its convenience for being accessibly located on the university campus and its similarity to several other lakes that are of interest for diverse limnology applications, including the study for growth patterns of "algal bloom". Nuisance algal bloom can impair the beneficial use of

---

8. Time for which the sensor has to be kept static to get an accurate measurement.

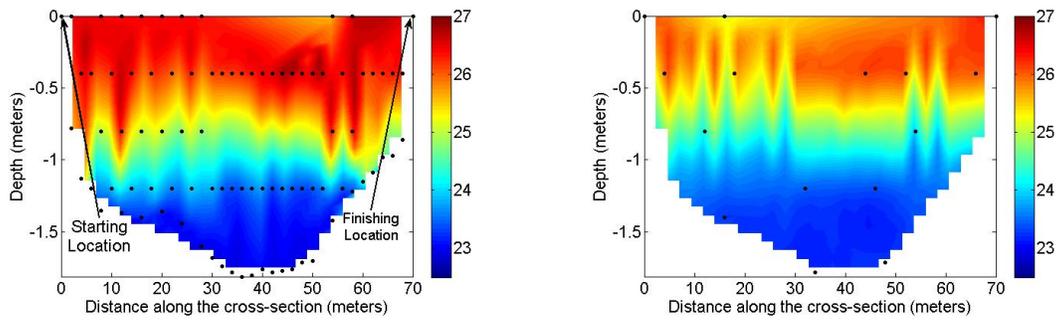

(a) Observed distribution during a raster scan on August 11

(b) Predicted distribution after observing at locations as output by e*MIP*

Figure 10: Distribution of temperature ($^o$C) at the little on UC Merced campus. Points represent observation locations during the corresponding experiment.

aquatic systems, by blocking sunlight to underwater vegetation, consuming oxygen in the water, and producing surface scum and odors. The growth pattern of algal bloom in a lake is dependent on the spatial and temporal dynamics of temperature, dissolved nutrients and light occurring in different layers of its environment. Thus, temperature is one of the critical parameter to observe in the lake environment as it controls several physical processes occurring in such low flow aquatic environments (in contrast to the San Joaquin river environment where there is considerable water flow).

The total width of the observed cross-section was 70 meters, with a maximum depth of up to 1.81 meters. Similarly to the San Joaquin deployment, we first learned the non-stationary GP model using the temperature data from one of the raster scans performed on August 10. Fig. 10a displays a typical surface distribution of temperature as observed during the raster scan at the lake. A total of 89 locations were observed during the raster scan. A set of 15 locations was selected out of these 89 locations (17%) using the e*MIP* algorithm with the starting and ending location on either end of the cross-section as displayed in Fig. 10a. This set of 15 observation locations was then observed the next day using NIMS as the robotic platform. Similar to San Joaquin deployment, we performed raster scans before and after our experiment to get a measure of ground truth for the temperature distribution. The predicted temperature, as computed after making the observations at the subset of locations selected using e*MIP*, is then compared with this ground truth. With a smaller dwelling time of 10 seconds (required for measuring temperature) and having to cover the entire length of the lake cross-section, the reduction in experimental time was 50% (when compared with the raster scan).

Fig. 10b displays the predicted distribution of temperature with points representing the observed locations as output by e*MIP*. Fig. 10a displays the distribution as observed using raster scan performed after the path planning experiment. The RMS error between the predicted distribution and the raster scan performed after the path planning experiment was 0.73 $^o$C. On the other hand, the RMS error between the predicted distribution and the raster scan performed before the path planning experiment was 0.82 $^o$C. The RMS error between the two raster scans performed before and after the path planning experiment, indicating the temporal variation in the environment, was 1.25 $^o$C. The low RMS error between the predicted distribution and the raster scans, in comparison with the temporal variation exhibited by the lake environment, indicates the effectiveness of our approach in the low-flow lake environment as well.

## 8.2 Experiments on Sensing Datasets

Several experiments were performed in simulation using real world sensing datasets to analyze the scaling of our algorithm for different approaches such as varying the experimental cost, exponential increase in budget split, varying the size of the cells of the spatial decomposition and comparison of several heuristics, among others. Three different datasets, collected from real world sensing applications, were used for these experiments. The first dataset consists of measurements of the temperature in Lake Fulmor, James Reserve (hereafter referred to as lake temperature dataset). Fig. 1b displays the aerial view of Lake Fulmor. A robotic boat, part of Networked Aquatic Microbial Observing System (NAMOS) (Dhariwal et al., 2006), was used to collect the surface temperature data around the lake, of width around 50 meters and length around 250 meters. As discussed earlier, understanding temperature distribution is of prime importance in limnology since it governs several physical phenomena occurring in the lake environment, including the growth of algal bloom.

The average speed of the boat was approximately 0.4 $m/s$. Half of the total measurements (218 different sensing locations) were used to learn a nonstationary Gaussian Process model by maximizing the marginal likelihood (Rasmussen & Williams, 2006), and the remaining measurements were used for experimentation. We divided the lake into 22 cells (except during the experiments for studying the effect of changing the size of the cell in spatial decomposition), with distance between adjacent cell approximately 21 meters. Based on the average speed, and motivated by a typical measurement duration of roughly 25 seconds, we set the experiment cost to be 10.5 meters (except during the experiment for understanding the effect of scaling the experimental cost).

As our second dataset, we used data from an existing deployment of 52 wireless sensor motes to learn the amount of temperature variability at the Intel Research Laboratory, Berkeley (hereafter referred to as Berkeley temperature dataset). These sensing locations lie within a bounding region of length 45 meters and width 40 meters. We divided the complete region into a uniform grid containing 20 equal sized cells, and determined the experimental cost to be 9 meters (approximate distance to travel between adjacent cells). We learned a GP model as discussed by Krause et al. (2006).

Finally, we explored the performance of our algorithm on a precipitation dataset collected from 167 regions of equal area, approximately 50 km apart, during the years 1949-1994. We followed the preprocessing and model learning described in the work by Krause et al. (2008). The large physical spread of the sensing regions makes this dataset unconventional for a mobile robot path planning application. To avoid this unrealistic scenario, we normalized the coordinates of the regions to lie within a bounding region of length 7 meters and width 9 meters, while keeping the actual sensing data observed at each location. We then divided the complete region into a uniform grid of 20 cells with experimental cost as 1.4 meters (approximate traveling distance between adjacent cells).

For each of the plots comparing the performance of our algorithm, x-axis represent the total cost of the path including both the traveling cost between the selected locations and the sensing cost at each selected location (translated into distance as discussed above). When comparing the computation effort as a measure of performance, in seconds, y-axis is drawn in logarithmic scale. The computation effort is for running the code implemented in Matlab on a 3.2 GHz dual processor core with 4 GB RAM. When comparing the collected reward as a measure of performance, y-axis represent the mutual information (submodular reward function) collected by making observations at the selected locations.

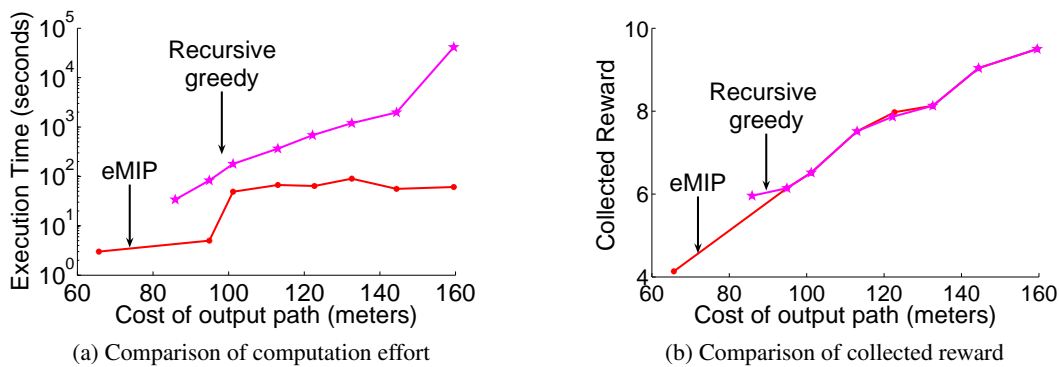

(a) Comparison of computation effort

(b) Comparison of collected reward

Figure 11: Comparison of e*MIP* and *recursive-greedy* on a subset of Berkeley temperature dataset with 23 sensing locations.

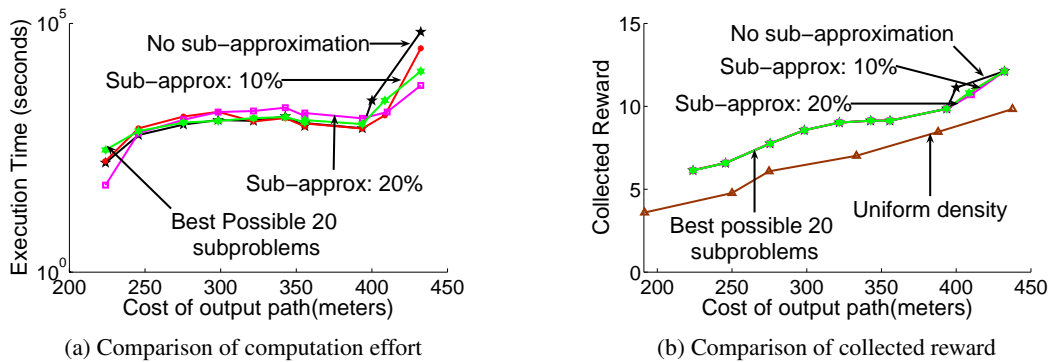

(a) Comparison of computation effort

(b) Comparison of collected reward

Figure 12: Comparison of computation effort and collected reward for several sub-approximation heuristics used to improve the running time of e*MIP* on lake temperature dataset. Significant improvement in execution time was observed, particularly for longer paths, without significant reduction in collected reward.

### 8.2.1 COMPARISON WITH RECURSIVE-GREEDY ALGORITHM:

To compare the performance of our approach with the *recursive-greedy* algorithm, as proposed by Chekuri et al., we selected a subset of 23 locations from the total of 52 locations from the Berkeley temperature dataset. A small subset of locations was selected since the running time of *recursive-greedy* is quasi-polynomial and was very large for the complete dataset. Fig. 11a and Fig. 11b display the comparison in the computation effort and collected reward on this smaller dataset for the two algorithms. As is evident from the plots, our approach provides significant improvement in running time (of several orders of magnitude at higher budget values) with (almost) the same collected reward. Since the recursive greedy algorithm is essentially a search procedure with greedily restricted search space, this result also indicates that an exhaustive search over all paths is intractable for even a small real world sensing problem. The sudden jump in execution time of e*MIP* in Fig. 11a at budget = 100 meters is due to an additional iteration step (*c.f.,* Line 7 in Algorithm 3) added due to the increase in the input budget constraint. Thereafter, additional increase in budget only results in increase in experimental budget. Since the *recursive*-e*SIP* computes efficiently for such a small problem, additional increase in experimental budget does not increase the computation effort significantly.

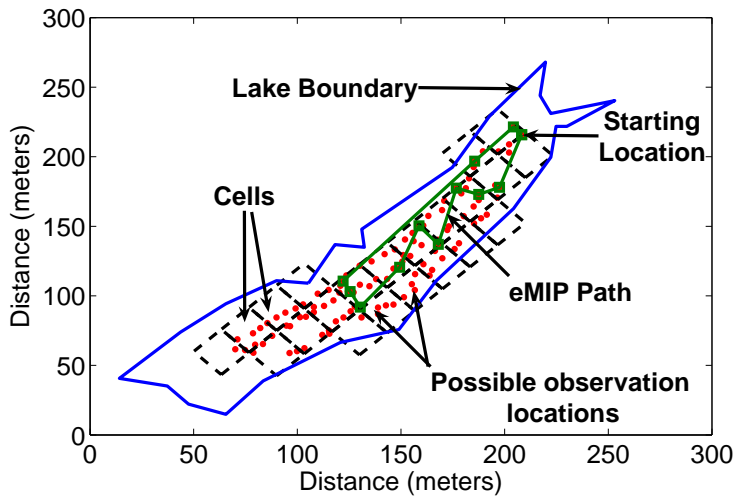

Figure 13: Illustration of a path selected using e*MIP* on lake temperature dataset.

### 8.2.2 COMPARISON WITH UNIFORM SAMPLE SPACING:

We compared the performance of e*MIP* with a simple uniform sample spacing algorithm, referred to as *Uniform density*. For the case of *Uniform density*, starting and finishing at given locations, we greedily select two observation locations within each of the nearest cells and compute the corresponding path cost and path reward. *Uniform density* algorithm will output the best possible path amongst all possible simple uniform sample spacing algorithms due to greedy observation selection within each cell. Fig. 12b, compares the collected reward for *Uniform density* with e*MIP* for the lake temperature dataset. Increased collected reward by e*MIP*, compared to *Uniform density*, empirically justifies the complexity of e*MIP*. Additionally, e*MIP* also provides a strong approximation guarantee which is not possible for any uniform sample spacing algorithm. Fig. 13 illustrates a path selected by e*MIP* for the lake temperature dataset, demonstrating that e*MIP* does not tend to cause uniform sample spacing. For a few of the traversed cells, there was no location selected for observation, while for others as many as three observation locations were selected from within the cell.

### 8.2.3 COMPARISON OF SUB-APPROXIMATION HEURISTICS:

Various sub-approximation heuristics discussed in Section 7 were compared empirically to analyze their utility in improving the execution time and the corresponding reduction in collected reward, if any. As is displayed in Fig. 12a that compares these heuristics for computation effort, each of these sub-approximation heuristic provides improvement in the execution time over the scenario when all branch and bound heuristics other than sub-approximation heuristics were used. The most improvement at higher values of input budget was observed when the lower bound was increased by a factor of $\alpha (= 1.2$ or 20%). Fig. 12b displays the corresponding comparison of these heuristics for collected reward. It was interesting to observe that none of the sub-approximation approaches resulted in considerable reduction in collected reward.

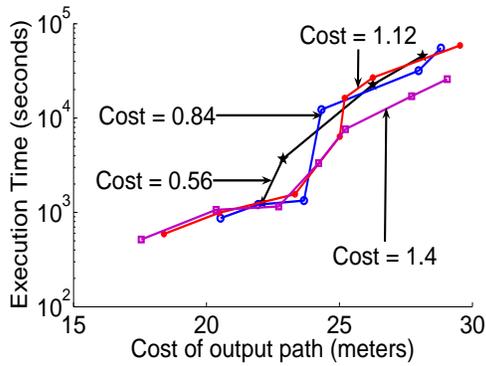

(a) Computation effort with variation in sensing cost using precipitation dataset

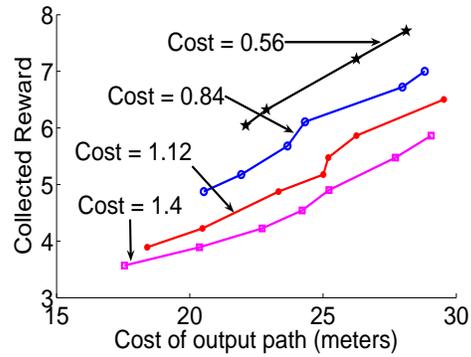

(b) Collected reward with variation in sensing cost using precipitation dataset

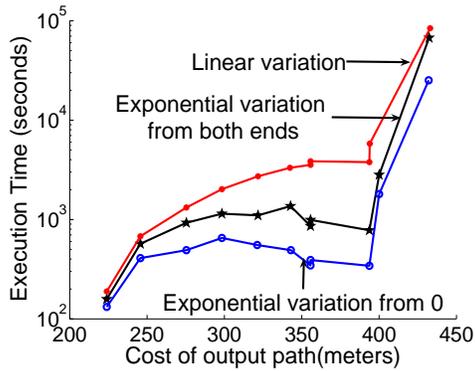

(c) Computation effort with variation in experimental budget split using lake temperature dataset

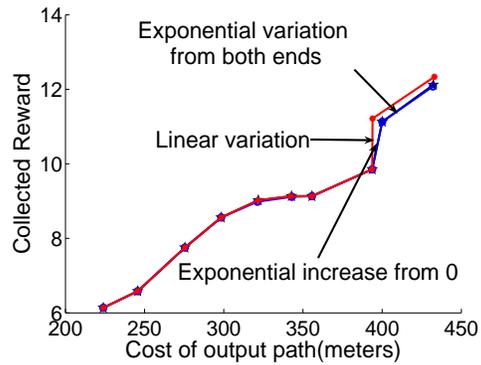

(d) Collected reward with variation in experimental budget split using lake temperature dataset

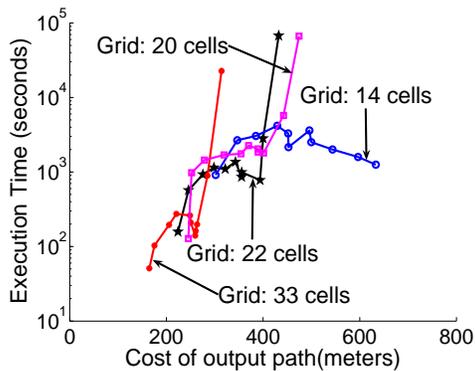

(e) Computation effort with variation in grid size for spatial decomposition using lake temperature dataset

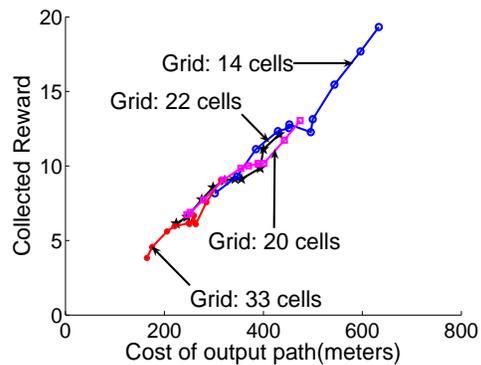

(f) Collected reward with variation in grid size for spatial decomposition using lake temperature dataset

Figure 14: Comparison of collected reward and computation effort with variation in several approaches used in e*MIP*.

### 8.2.4 Variation in Sensing Cost:

Fig. 14a and Fig. 14b compare the computation effort and collected reward as the sensing cost is varied for the precipitation dataset. With the reduction in experimental cost, more locations were observed for the same total input budget resulting in increased collected reward. However, for each of the experiments, the computation effort was approximately the same. Due to the diversity in environmental applications, the sensing cost will depend on the sensors (settling time) and the scale of dynamics occurring in the observed phenomena. This experiment indicates that e*MIP* can be used over a diverse range of sensing costs, as per the demands of diverse environmental applications.

### 8.2.5 Variation in Experimental Budget Split:

As discussed in Section 6, the strategy of exponentially increasing the experimental budget split results in an increased additional path length required to guarantee the approximation factor for the collected reward. We performed several experiments with the available datasets to analyze the empirical performance of increasing the budget splits exponentially. Fig. 14c and Fig. 14d compares the computation effort and collected reward for linear increase, one sided exponential variation from 0 and two-sided exponential variation from both 0 and budget $B$ for the lake temperature dataset. Since a smaller number of budget splits are considered in *recursive*-e*SIP* in the case of an exponential increase, the computation effort will be smaller as compared to the linear increase in the budget splits. Interestingly, there was very small reduction in collected reward, for only a few budget values, when the exponential increase was employed. Hence, even though the theoretical approximation guarantee with exponential increase in experimental budget is weaker, empirically the collected reward for both the linear and exponential increase in budget splits was found to be comparable over a wide range of input budgets.

### 8.2.6 Analysis of Spatial Decomposition:

As discussed in Section 6, the conversion of an SD-MIPP solution (a cell path) into a solution for MIPP (a path over observation locations) will result in additional path length exceeding the input budget $B$. This additional path length will depend on the size of the cell (or size of the grid covering the complete spatial domain) in SD-MIPP problem and will result in trade-off with the computation effort. Variation in grid-size will result in corresponding variation in the traveling cost between the neighboring cells. This will result in an opportunity to travel more cells for a denser grid with the same input budget constraint. However, to keep the experimental cost constant across the varying grid size (since the experiment cost only depends on observed phenomena and is independent of the spatial decomposition), it was scaled accordingly, in proportion to the traveling cost between the neighboring cells. Fig. 14f compares the collected reward for varying grid sizes on the lake temperature dataset, changing the grid size from 14 to 33 cells. It is interesting to observe that such a change in grid size had (almost) negligible effect on the collected reward. On the other hand, such increase in grid density resulted in a larger number of cells over which path planning is to be performed thus leading to increased computation effort for the same input budget. The comparison of the computation effort for the varying grid size is displayed in Fig. 14e. Note the drastic increase in computation time as the grid discretization is made finer.

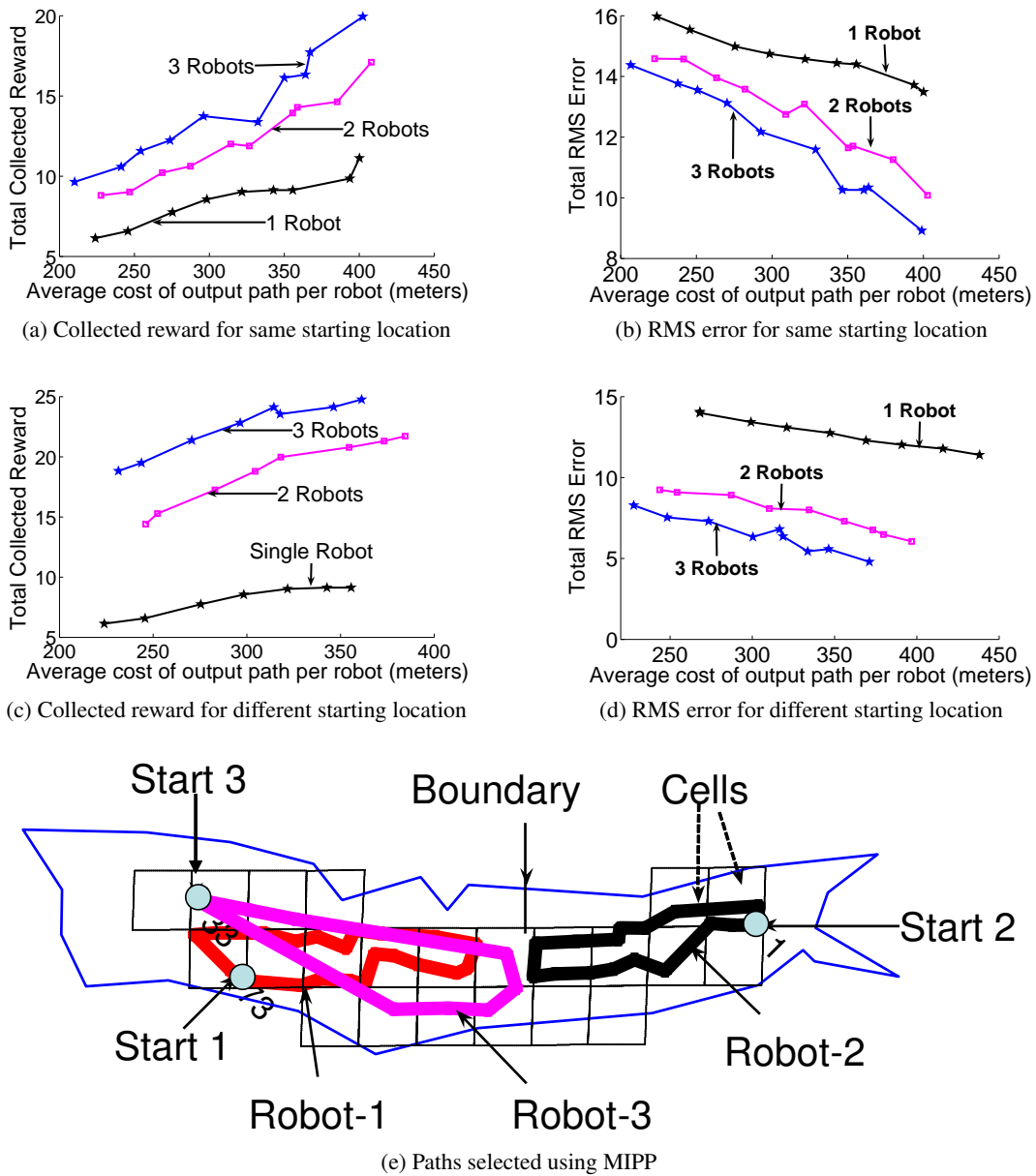

Figure 15: Analysis of experiments performed for multiple robots with different (optimized) starting location using the lake temperature dataset.

### 8.2.7 MULTI-ROBOT EXPERIMENTS

We evaluated the performance of our e*MIP* multi-robot algorithm in simulation using several sensing datasets. Fig. 15 displays the empirical analysis of several experiments using the lake temperature dataset. The first experiment was performed with each robot starting from the same starting location. Fig. 15a and Fig. 15b display the collected reward and root mean square (RMS) error when the number of robots were varied from one to three. Due to the *sequential-allocation* ap-proach (wherein we remove the locations that are already selected before selecting the locations for the next robot) and "information never hurts" principle, collected reward increases as the number of robots were increased and hence the corresponding root mean square error for prediction at the unobserved locations gets reduced. However, the incremental change in performance from one to two robots was larger than the incremental change from two to three robots, which is expected from the submodularity (diminishing returns) property of mutual information.

Fig. 15c and Fig. 15d display the collected reward and RMS error when a different starting location is chosen for each robot. In this scenario, a set of four starting locations is pre-determined with each location at one end of the lake (see for reference Fig. 15e where three of the four starting locations are marked). The starting location for each of the three robots was selected greedily based on the collected information. With a different starting location selected on the opposite end of the lake for the second robot, the incremental change in collected reward (and corresponding decrease in root mean square error) as the number of robots was increased from one to two is much higher than the corresponding change when the same starting location was chosen for the second robot as well. However, similar to the scenario with same starting location, the incremental change as the number of robots was increased from one to two is higher as compared to when the number of robots was increased from two to three (due to submodularity of mutual information). Fig. 15e illustrates the selected paths for each of the three robots as selected using e*MIP*.

## 9. Related Work

There is a large body of related work both in the theory of path planning and its applications. Approximation algorithms have been proposed for several related problems. Variants of path planning have been studied in the field of Operations Research as the Traveling Salesman Problem (TSP) or the Vehicle Routing Problem (VRP). In robotics, several path planning approaches have been studied for applications such as Simultaneous Localization and Mapping (SLAM) and search and exploration. In sensor networks and geostatistics, a closely related work studies optimal placement of static sensors modeling the phenomenon as *Gaussian Processes*. Several adaptive sampling approaches have been studied to decide on the subset of locations to observe in order to understand the phenomenon dynamics effectively. In addition, similar approaches are explored for planning paths for mobile robots acting as data mules, collecting data sampled by the network of static sensors.

### 9.1 Operations Research

An interesting special case of the MIPP problem is given in the case where each node has a fixed reward, and the goal is to find a path that maximizes the sum of these rewards (*Traveling Salesman Problem with Profits*, TSPP, Feillet et al., 2005). Such a sum of rewards is a *modular* (additive) function, which is a special case of submodular functions. A subcategory of TSPP is an optimization problem defined to maximize the collected reward while keeping the associated cost less than some given budget $B$. This was studied as *Orienteering Problem* (OP) or *selective TSP* (Laporte & Martello, 1990), or *Maximum Collection Problem* (Kataoka & Morito, 1988) in the literature. The additivity assumption made in the orienteering problem is very unrealistic in our informative path planning setting, as it assumes that the information provided by adjacent locations is *independent*, whereas we would typically expect a strong amount of correlation. In fact, if the observations were all independent, there would be no point in selecting observations for spatial prediction. In this paper, we hence study the more general orienteering problem with submodular reward functions, proposed earlier as *Submodular Orienteering Problem* (Chekuri & Pal, 2005).

### 9.1.1 MULTIPLE-PATH EXTENSIONS:

The extension of TSPP to multiple paths was studied as Vehicle Routing Problem with Profits (VRPP) in the literature. Like TSPP, several variants of VRPP have been previously considered. The Prize Collecting VRP (PCVRP) (Tang & Wang, 2006) is a class of VRPP where the objective is to determine a subset of all customers to visit so as to minimize the total distance traveled, minimize the vehicles used and maximize the collected reward. The multi-robot version of the OP (in the case of additive reward functions) was studied as the *Team Orienteering Problem* by I-Ming et al. (1996) and *Multiple Tour Maximum Collection Problem* by Butt and Ryan (1999).

### 9.1.2 KNOWN APPROXIMATIONS FOR THE ORIENTEERING PROBLEM:

The OP is known to be **NP**-hard (Golden et al., 1987). Several versions of the OP studied in the literature can be classified into those for which the starting (and the finishing) location (root) is pre-specified or not. For the case of the unrooted OP (when no starting location is specified), the approximation guarantees known for *Prize Collecting TSP* and k-TSP can be easily extended (Johnson et al., 2000). There are several constant factor approximations known for the PC-TSP and k-TSP problems with the best one being a 2 approximation (Garg, 2005). However the same extension does not apply for the rooted version of the problem as the best path for the unrooted version may not contain the root and may be far away from the root thus leading to violation of the budget constraint.

For the rooted OP, Arkin et al. (1998) gave a $(2 + \epsilon)$ approximation for the OP in geometric settings. Blum et al. (2003) gave the first constant factor approximation for the rooted OP in general undirected graphs. They also extended their algorithm for multi-path OP. The running time of their algorithm, though polynomial, is very large (more specifically, $\mathcal{O}(\pi n^5 \log(\frac{1}{\epsilon}))$ where $\pi$ is the total reward in the path). Recently Chekuri et al. (2008) gave a polynomial time algorithm for the OP in undirected graphs with an improved approximation guarantee of $(2 + \epsilon)$. Our problem formulation with specified starting location ($s$) and finishing location ($t$) falls under the category of rooted OP with submodular (non-additive) reward function.

Another classification of OP can be done based on the symmetry of the space of the possible locations. All of the above approximation guarantees hold true on symmetric spaces (undirected graphs). Obtaining good approximation algorithm for the directed (asymmetric) orienteering problem was stated as an open problem by Blum et al. (2003). Chekuri and Pal (2005) gave the first approximation algorithm with $\mathcal{O}(\log n)$ guarantee that runs in quasi-polynomial running time. The running time was recently improved independently by two different works (Chekuri et al., 2008; Nagarajan & Ravi, 2007), each proposing a poly-time approximation algorithm providing an approximation guarantee of $\mathcal{O}(\log^2 n)$, though using different approaches. The metric space conversion procedure used during our spatial decomposition approach limits e*MIP* to symmetric spaces only.

### 9.1.3 SEQUENTIAL ALLOCATION:

Blum et al. (2003) proposed a sequential allocation approach to extend algorithms for single-robot orienteering to the multiple robot setting, but only for the special case of additive (modular) reward functions. In this paper, we generalize their result to submodular reward functions. After the initial version of this paper was published (Singh et al., 2007b), we realized that our *sequential-allocation* procedure is an instance of maximizing a submodular function subject to a matroid constraint (Calinescu et al., 2007). We can define a partition matroid on the disjoint union $\mathcal{M} = \mathcal{M}_1 \cup \cdots \cup \mathcal{M}_k$ of $k$ ground sets $\mathcal{M}_i$, one for each robot. Each set $\mathcal{M}_i$ contains all feasible paths for robot $i$. The collection $\mathcal{I} \subseteq 2^{\mathcal{M}}$ of all subsets $\mathcal{P} \in \mathcal{I}$ such that $|\mathcal{P} \cap \mathcal{M}_i| \leq 1$ (i.e. each $\mathcal{P}$ corresponds to a collection of paths, with the constraint that we can pick at most one from each set $\mathcal{M}_i$) forms independent sets of the partition matroid. Hence, the problem of finding a collection of maximally informative paths is the problem of finding an independent set of a matroid maximizing a submodular function. Current work in progress by Goundan and Schulz (2008) provides general results on the performance of a sequential allocation procedure in such a setting, which can be used to prove the same sequential allocation results originally presented by Singh et al. (2007b).

## 9.2 Robotic Applications

There is considerable work in path planning in the robotics community for several applications, including *simultaneous localization and mapping* (SLAM) and search and exploration. Several different approaches have been studied for each of these applications, including auction based algorithms, data-adaptive approaches and information gain based algorithms.

### 9.2.1 Simultaneous Localization and Mapping:

The goal of Simultaneous Localization And Mapping (SLAM) is to build maps of an environment by performing an exploration of the environment with an objective to estimate the robot position and world features simultaneously. Several approaches optimizing different objective functions had been proposed to perform path planning for SLAM. Bourgault et al. (2002) proposed an exploration framework using an occupancy grid (OG) environment model (performing spatial decomposition of the observed environment) with an objective to maximize mutual information over the OG map. Stachniss et al. (2005) developed a greedy algorithm for selecting the next location to visit to maximize information gain about the map.

In contrast to such approaches, Sim and Roy (2005) attempted to optimize the entire trajectory, not just the next step, but their algorithm introduces some approximations without theoretical bounds. Simmons et al. (2000) proposed a distributed approach for exploration and mapping with multiple robots by minimizing the overlap in information gain amongst multiple robots. They provided quantitative results from simulation but did not provide any theoretical bounds for their approach. There is little work in SLAM setting with an upper bound on the total cost of the path. In addition, we are not aware of any approaches to SLAM which carry approximation guarantees for either the single or multi-robot cases. An interesting direction for future work would be to analyze the applicability of our approach to the SLAM setting.

### 9.2.2 Search and Exploration:

The search and exploration application involves path planning for a robot with the goal of searching for a moving target(s) in a given environment, e.g. target surveillance in security applications and patient tracking in health care domain. Performing path planning using stochastic inference provides advantage of robustness to sensing and motion uncertainty though with an added complexity of computational intractability. Roy and Earnest (2006) proposed an approach to effectively compute the trajectories for target tracking based on maximizing mutual information (evaluated using the change in variance of the probability distribution). They used a particle filter approach, performing clustering over the particles followed by path planning over these clusters. Lau et al. (2006) formulated the target tracking in indoor environments as a generalization of an **NP**-complete optimal searcher path (OSP) problem (Trummel & Weisinger, 1986). They sought to optimize the probability of detection within a given time horizon while accounting for the undetected target probability that is a function of previously visited locations during the search. They used several branch and bound approaches to speed up the search process. The objective of maximizing information gain subject to the budget constraints on the path cost makes e*MIP* a suitable candidate for performing path planning for such problems.

Ryan (2008) used an approach of partitioning the search space into subgraphs for multi-robot path planning. We take a conceptually similar approach, also reducing the search space by decomposing the space into regions and then performing path planning over those regions. However, we address more complex utility functions, such as quantifying the informativeness of visited locations and are not limited to specific graph structures such as stacks, halls, cliques, rings as is the case in the work of Ryan (2008). Recently, Thompson and Wettergreen (2008) used our e*MIP* algorithm for near-term path planning while performing autonomous exploration of surficial units at Amboy Crater in Mojave desert, California.

### 9.2.3 PLANNING SYSTEMS AND APPLICATIONS:

Certain applications in robotic path planning used plan graphs (Blum & Furst, 1997) to compute an estimate of the resources and time required to achieve goals from states encountered in the search process. In the case of over-subscription planning problem – wherein only a subset of goals that can be accomplished within the limited time or resources available for the planning system, the work by Smith (2004) used an orienteering heuristic to provide an ordered set of goals to be considered by the planner. Briel et al. (2004) proposed several heuristics for efficiently solving the over-subscription planning problem. However, in each of the earlier proposed heuristics, the reward function considered is modular (additive). e*MIP* can be used to efficiently solve the over-subscription planning problem in the submodular setting with strong approximation guarantees.

### 9.3 Sensor Networks

Phenomenon modeling to decide on the optimal placement of a set of static sensors is well studied in the sensor networks and geostatistics communities. Gaussian Process models for spatial phenomena had been studied extensively (Cressie, 1991). Guestrin et al. (2005) proved that, in the case of phenomena governed by Gaussian Process models, selecting the placement of sensors greedily based on mutual information is near-optimal. Krause et al. (2006) extended this work to include communication cost between sensors while optimizing the sensor placement. In the communication constrained setting, similar to the path planning problem considered in this paper, the greedy algorithm performs badly, and more involved algorithms have to be developed. Batalin et al. (2004) showed that combining the static and mobile sensing devices, even in a simple scenario, can result in significant improvement in sensing performance. In such a scenario, where a combination of static and mobile sensing devices are available, several approaches for optimal placement of static sensors can be combined with e*MIP* to observe a given phenomenon efficiently.

### 9.3.1 DATA COLLECTION FROM A SENSOR NETWORK:

A different scenario where a mobile robot can be combined with a network of static sensors is to improve the lifetime of the sensor network by performing the tours for collecting the data sampled by the static network. Somasundara et al. (2007) showed that the problem of collecting the data when the environment shows both spatial and temporal dynamics is **NP**-complete and provided an integer linear programming formulation for the same. They compared the performance of several heuristics in simulation for both single and multi-robot scenario. Meliou et al. (2007) proposed a nonmyopic approach for the application of data gathering tours using an algorithm for submodular orienteering (SOP) as a black box. They provided strong approximation guarantees and extensive empirical evaluation that indicates the applicability of their approach for such applications. In this setting, e*MIP* can be used as an orienteering algorithm to provide a better approximation guarantee in addition to improved running time.

### 9.3.2 ADAPTIVE SAMPLING FOR ENVIRONMENTAL APPLICATIONS:

Recent advances in robotics have opened up opportunities for high fidelity monitoring of dynamic environmental sensing applications. Rahimi et al. (2004) explored several policies for adaptively sampling the environment. Singh et al. (2006) proposed a multiscale adaptive sampling approach with uniformly sampling the environment in the first stage followed by sampling at locations in order to minimize the mean square error the most. They also extended their approach for multiple robots, although without providing any theoretical bounds. Using several in-field experiments as well as simulations using real world sensing datasets, we demonstrate here that several such environmental phenomenon can be effectively sampled adaptively using e*MIP*.

## 10. Conclusions and Future Work

In this paper, we presented e*SIP*, an approximation algorithm for efficient planning of informative paths. e*SIP* near-optimally solves the **NP**-hard problem of maximizing the collected information with an upper bound on *path-cost*. Our e*SIP* algorithm builds on the *recursive-greedy* algorithm of Chekuri and Pal (2005). e*SIP* preserves the approximation guarantees of *recursive-greedy*, while overcoming its computational intractability through *spatial-decomposition* and several *branch and bound* approaches. We also presented a general approach, *sequential-allocation*, which extends any single-robot algorithm, such as e*SIP*, to the multiple-robot setting while providing a provably strong approximation guarantee.

We also provide extensive empirical evaluation to demonstrate the effectiveness of our approach for real world sensing applications. We performed several in-field experiments for two important environmental sensing applications – lake monitoring (at a small lake at UC Merced campus) and river monitoring (at San Joaquin river, California). The Networked Info Mechanical System (NIMS) was used as the robotic system for performing path planning during each of these deployments to demonstrate the practicality of our algorithm. We also performed extensive simulation experiments using several real world sensor network data sets. With global climate change and corresponding impetus on sustainable practices, we expect that such efficient path planning approaches can help address the challenge of monitoring environment-related activities effectively.

In the future, we plan to explore the applicability of our algorithm in other application domains such as SLAM and search and rescue. We plan to work towards understanding the limitations of learning a static GP model in real world scenarios, and extend our approach for online model adaptation.

## Acknowledgments

We would like to thank Maxim Batalin for helpful discussions, Bin Zhang for providing the lake data set and Michael Stealey, Henry Pai and Victor Chen for help during the river and lake deployment. This work was partially supported by NSF Grants No. CNS-0509383, CNS-0625518, CNS-0331481, ANI-00331481, CCR-0120778, ECCS-0725441, ONR MURI W911NF0710287 and a gift from Intel. Carlos Guestrin was partly supported by an Alfred P. Sloan Fellowship and an IBM Faculty Fellowship. Andreas Krause was partially supported by a Microsoft Research Graduate Fellowship.

## APPENDIX

**Theorem-1.** *Let $\eta$ be the approximation guarantee for the single path instance of the informative path planning problem. Then our sequential-allocation algorithm achieves an approximation guarantee of $(1+\eta)$ for the MIPP problem. In the special case, where all robots have the same starting $(s_i = s_j, \forall i,j)$ and finishing locations $(t_i = t_j, \forall i,j)$, the approximation guarantee improves to $1/(1 - \exp{(-1/\eta)}) \leq 1 + \eta$.*

*Proof of Theorem 1.* For the case when all the robots start and finish at the same location, let $\Pi$ be the total reward collected by the optimal solution. Additionally, define $\Pi_i$ to be the difference between the reward collected by the optimal solution, and by the approximation algorithm, at the end of stage $i$. Hence, $\Pi_0 = \Pi$.

Let $\mathcal{A}_i = \mathcal{P}_1 \cup \cdots \cup \mathcal{P}_i$ be the nodes selected by the approximation algorithm up to stage $i$ ($\mathcal{A}_0 = \emptyset$), and let $\mathcal{P}^* = \{\mathcal{P}_1^*, \ldots, \mathcal{P}_k^*\}$ denote the collection of paths chosen in the optimal solution. Consider the residual reward $f_{\mathcal{A}_i}$. We find $f_{\mathcal{A}_i}(\mathcal{P}^*) = f(\mathcal{A}_i \cup \mathcal{P}^*) - f(\mathcal{A}_i) \geq f(\mathcal{P}^*) - f(\mathcal{A}_i) = \Pi_i$ due to monotonicity of $f$. If there were no path $\mathcal{P}_j^*$ with $f_{\mathcal{A}_i}(\mathcal{P}_j^*) \geq \frac{1}{k}\Pi_i$, then $\sum_j f_{\mathcal{A}_i}(\mathcal{P}_j^*) < \Pi_i = f_{\mathcal{A}_i}(\mathcal{P}^*)$, contradicting the monotonic submodularity of $f_{\mathcal{A}_i}$. Hence there is such a path $\mathcal{P}_j^*$ with $f_{\mathcal{A}_i}(\mathcal{P}_j^*) \geq \frac{1}{k}\Pi_i$, and thus the approximation algorithm is guaranteed to find a path $\mathcal{P}_i$ such that $f_{\mathcal{A}_i}(\mathcal{P}_i) \geq \frac{1}{\eta k}\Pi_i$.

The difference in the reward collected by the optimal solution and the reward collected by Algorithm 1 after stage $i + 1$ is at most:

$$\Pi_{i+1} \leq (1 - 1/\eta k)\Pi_i,$$
$$\leq (1 - 1/\eta k)^{i+1}\Pi.$$

Thus after $k$ stages, the difference in the reward is bounded by $\Pi_k \leq (1-1/\eta k)^k \Pi \leq \exp{(-1/\eta)}\Pi$. Hence, the reward collect by Algorithm 1 is at least $(1 - \exp{(-1/\eta)})$ times the optimal reward, resulting in approximation factor of $1/(1 - \exp{(-1/\eta)})$.

For the case when each robot has different starting and finishing location, let $\mathcal{P}_i^*$ be the set of nodes visited by the optimal path at stage $i$. Let $\mathcal{O}_i$ be the set of nodes visited by the optimal path until stage $i$, i.e., $\mathcal{O}_i = \cup_{j=1}^i \mathcal{P}_j^*$, with $\mathcal{O}_0 = \emptyset$ and $\mathcal{O}_1 = \mathcal{P}_1^*$. The reward collected by the approximation algorithm at stage $i$ can be bounded as:

$$f_{\mathcal{A}_{i-1}}(\mathcal{P}_i) \geq 1/\eta(f_{\mathcal{A}_{i-1}}(\mathcal{P}_i^*)).$$

After $k$ stages, the total collected reward can be given as:

$$\sum_{i=1}^{k} f_{\mathcal{A}_{i-1}}(\mathcal{P}_i) \geq 1/\eta(\sum_{i=1}^{k} f_{\mathcal{A}_{i-1}}(\mathcal{P}_i^*)). \tag{4}$$

Since the left hand side is a telescopic sum, we get:

$$\sum_{i=1}^{k} f_{\mathcal{A}_{i-1}}(\mathcal{P}_i) = f(\cup_{i=1}^{k} \mathcal{P}_i) = f(\mathcal{A}_k). \tag{5}$$

On the right hand side (RHS):

$$R.H.S. = 1/\eta(\sum_{i=1}^{k} f_{\mathcal{A}_{i-1}}(\mathcal{P}_i^*)),$$

$$= 1/\eta(\sum_{i=1}^{k}(f(\mathcal{P}_i^* \cup \mathcal{A}_{i-1}) - f(\mathcal{A}_{i-1}))).$$

Adding $\mathcal{O}_{i-1}$ to both the terms and using the submodularity property, we get

$$R.H.S. \geq 1/\eta(\sum_{i=1}^{k}(f(\mathcal{O}_i \cup \mathcal{A}_{i-1}) - f(\mathcal{O}_{i-1} \cup \mathcal{A}_{i-1}))),$$

$$= 1/\eta \left[ f(\mathcal{O}_1) - 0 + f(\mathcal{O}_2 \cup \mathcal{A}_1) - f(\mathcal{O}_1 \cup \mathcal{A}_1) + \cdots + f(\mathcal{O}_k \cup \mathcal{A}_{k-1}) - f(\mathcal{O}_{k-1} \cup \mathcal{A}_{k-1}) \right].$$

Rearranging the terms, we get:

$$R.H.S. \geq 1/\eta \left[ f(\mathcal{O}_k \cup \mathcal{A}_{k-1}) - \sum_{i=1}^{k-1}(f(\mathcal{O}_i \cup \mathcal{A}_i) - f(\mathcal{O}_i \cup \mathcal{A}_{i-1})) \right].$$

Using the monotonicity ($f(\mathcal{O}_k \cup \mathcal{A}_{k-1}) \geq f(\mathcal{O}_k)$) and submodularity of $f$ ( $f(\mathcal{O}_i \cup \mathcal{A}_i) - f(\mathcal{O}_i \cup \mathcal{A}_{i-1}) \leq f(\mathcal{A}_i) - f(\mathcal{A}_{i-1})$), we get

$$R.H.S. \geq 1/\eta \left[ f(\mathcal{O}_k) - \sum_{i=1}^{k-1}(f(\mathcal{A}_i) - f(\mathcal{A}_{i-1})) \right],$$

$$= 1/\eta \left[ f(\mathcal{O}_k) - f(\mathcal{A}_{k-1}) \right].$$

Using the monotonicity ($f(\mathcal{A}_k) \geq f(\mathcal{A}_{k-1})$), we get

$$R.H.S. \geq 1/\eta \left[ f(\mathcal{O}_k) - f(\mathcal{A}_k) \right]. \tag{6}$$

Substituting Equation (5) and (6) into Equation (4), we get:

$$f(\mathcal{A}_k) \geq 1/\eta \left[ f(\mathcal{O}_k) - f(\mathcal{A}_k) \right],$$

and thus:

$$f(\mathcal{A}_k) \geq 1/(\eta + 1)f(\mathcal{O}_k).$$

resulting in an approximation guarantee of $(1 + \eta)$.

$\square$

The above theorem and proof is inspired by the proof of multi-path orienteering provided by Blum et al. (2003).

**Lemma 3.** *Let* $\mathcal{P}^* = (s = v_0, v_1, \ldots, v_l = t)$ *be an optimal* $s$-$t$-*path solution to MIPP, constrained by budget* $B$*. Then there exists a corresponding SD-MIPP path* $\mathcal{P}_C^* = (\mathcal{C}_s = \mathcal{C}_{i_1}, \ldots, \mathcal{C}_{i_n} = \mathcal{C}_t)$*, traversing through locations* $\mathcal{A}_{i_1} \cup \cdots \cup \mathcal{A}_{i_l}$*, with budget* $\widetilde{\mathcal{B}}$ *of at most* $2\sqrt{2}B + 4L$ *collecting the same information.*

*Proof of Lemma 3.* Let $\mathcal{P}^*$ be the optimal path for MIPP, constrained by budget B. We need to ensure that when MIPP is transformed into SD-MIPP, with $\mathcal{P}_C^*$ as the corresponding optimal solution, we have enough budget such that $\mathcal{P}_C^*$ is feasible in the new problem domain. To recall, for the new problem domain, SD-MIPP, traveling to a new cell costs $L$ (distance between the centroids of adjacent cells), irrespective of the sensing location within the cell.

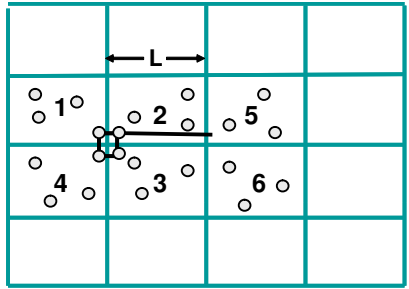

Figure 16: Illustration for the increased budget requirement for SD-MIPP.

For the corresponding SD-MIPP, an optimal path may just make 4 experiments in 4 different cells (Cells 1,2,3 and 4 in Fig. 16) sharing a common vertex, with each sensing location in different cell close to the common vertex, while only requiring an infinitesimally small traveling cost. Increasing the budget by $4L$ accounts for this case. Furthermore, by paying only an additional cost $L$ for traveling between the two corners of an edge of a cell, $\mathcal{P}_C^*$ can make experiments at 2 new cells (Cells 5,6 in Fig. 16. Thus, the total number of cells visited by the $\mathcal{P}_C^*$ is upper bounded by $2(B/L) + 4$. Hence, a budget of $2B + 4L$ suffices to render $\mathcal{P}_C^*$ a feasible SD-MIPP solution. Now to convert MIPP from the two-dimensional Euclidean distance into the corresponding $L_1$ distance, the budget needs to be increased to $\sqrt{2}B$ to ensure that $\mathcal{P}^*$ is feasible in the $L_1$ metric. Accounting for the conversion from Euclidean distance into $L_1$, the total budget $\tilde{B}$ required for SD-MIPP, to ensure the feasibility of the optimal solution in MIPP, is upper bounded by $2\sqrt{2}B + 4L$. $\square$

**Lemma 4.** *Let* $\mathcal{P}_C^* = (\mathcal{C}_s = \mathcal{C}_1, \ldots, \mathcal{C}_k = \mathcal{C}_t)$ *be an optimal solution for single robot instance of SD-MIPP, constrained by budget* $\widetilde{\mathcal{B}}$*, where an optimal set of locations are selected within each visited cell* $\mathcal{C}_j$*. Let* $\widehat{\mathcal{P}}$ *be the solution returned for* $e$*SIP. Then* $\mathrm{I}(\widehat{\mathcal{P}}) \geq \frac{1-1/e}{1+\log_2 k} \mathrm{I}(\mathcal{P}_C^*)$*.*

*Proof of Lemma 4.* We will prove this by induction on the length $n$ of the optimal path. Let $F_g(= (1 - 1/e))$ be the constant factor due to the greedy selection of sensing locations within each cell.

Also assume $\widetilde{\mathcal{B}}$ be the budget constraint for SD-MIPP problem. For the case $n = 1$, $iter = 0$ and Algorithm 4 will select the greedy subset of nodes from the set $\mathcal{C}_s = \mathcal{C}_t$. This will give an approximation guarantee of $F_g$ (Krause et al., 2008) compared to the optimal set of the same number of observations selected in this cell (and hence of the information obtained by the optimal SD-MIPP path visiting only this cell).

Now, assuming the induction hypothesis holds for $n = k/2$, we get:

$$\mathrm{I}_X(\mathcal{P}) \geq \frac{F_g}{(1 + \log(k/2))}\, \mathrm{I}_X(\mathcal{P}^*),$$
$$\geq \frac{F_g}{\log k}\, \mathrm{I}_X(\mathcal{P}^*).$$

This will hold true for traveling budget of $\widetilde{\mathcal{B}}_{k/2}$ and experimental budget up to $\widetilde{\mathcal{B}} - \widetilde{\mathcal{B}}_{k/2}$. Let us now analyze the case $n = k$. Let $P_1^*$ be the optimal path from $\mathcal{C}_s$ to $\mathcal{C}_{k/2}$ constrained by budget $B'$. Since we increase the experimental budget split linearly, $B'$ will vary from $0$ to $\widetilde{\mathcal{B}} - \widetilde{\mathcal{B}}_k$, where $\widetilde{\mathcal{B}}_k$ is the traveling cost for visiting $k$ cells. Since this cost will be less than $\widetilde{\mathcal{B}} - \widetilde{\mathcal{B}}_{k/2}$, using the induction hypothesis,

$$\mathrm{I}_X(\mathcal{P}_1) \geq \frac{F_g}{\log k}\, \mathrm{I}_X(\mathcal{P}_1^*). \tag{7}$$

Similarly, with $X' = X \cup \mathcal{P}_1$ following approximation guarantee holds true for $\mathcal{P}_2$:

$$\mathrm{I}_{X'}(\mathcal{P}_2) \geq \frac{F_g}{\log k}\, \mathrm{I}_{X'}(\mathcal{P}_2^*). \tag{8}$$

By definition of our submodular function:

$$\mathrm{I}_{X'}(\mathcal{P}_2^*) = \mathrm{I}(\mathcal{P}_2^* \cup \mathcal{P}_1 \cup X) - \mathrm{I}(\mathcal{P}_1 \cup X),$$
$$= \mathrm{I}_X(\mathcal{P}_1 \cup \mathcal{P}_2^*) - \mathrm{I}_X(\mathcal{P}_1).$$

Substituting in (8), we get

$$\mathrm{I}_{X'}(\mathcal{P}_2) \geq \frac{F_g}{\log k}(\mathrm{I}_X(\mathcal{P}_1 \cup \mathcal{P}_2^*) - \mathrm{I}_X(\mathcal{P}_1)).$$

Using monotonicity of I,

$$\mathrm{I}_{X'}(\mathcal{P}_2) \geq \frac{F_g}{\log k}(\mathrm{I}_X(\mathcal{P}_2^*) - \mathrm{I}_X(\mathcal{P})).$$

Adding this to (7), we finally get:

$$\mathrm{I}_X(\mathcal{P}) \geq \frac{F_g}{\log k}(\mathrm{I}_X(\mathcal{P}_1^*) + \mathrm{I}_X(\mathcal{P}_2^*) - \mathrm{I}_X(\mathcal{P})),$$
$$(F_g + \log k)\, \mathrm{I}_X(\mathcal{P}) \geq F_g(\mathrm{I}_X(\mathcal{P}_1^*) + \mathrm{I}_X(\mathcal{P}_2^*)),$$
$$(1 + \log k)\, \mathrm{I}_X(\mathcal{P}) \geq F_g(\mathrm{I}_X(\mathcal{P}_1^*) + \mathrm{I}_X(\mathcal{P}_2^*)).$$

Since $\mathrm{I}_X$ is a submodular function,

$$(1 + \log k)\, \mathrm{I}_X(\mathcal{P}) \geq F_g(\mathrm{I}_X(\mathcal{P}^*)),$$
$$\mathrm{I}_X(\mathcal{P}) \geq \frac{F_g}{1 + \log k}(\mathrm{I}_X(\mathcal{P}^*)).$$

□

The above proof is inspired by the analysis of the recursive greedy algorithm for submodular orienteering proposed by Chekuri and Pal (2005).

In the case of exponential budget splits, the budget needs to be increased, albeit sub-linearly:

**Lemma 7.** *Let* $\mathcal{P}_C^* = (\mathcal{C}_s = \mathcal{C}_{i_1}, \ldots, \mathcal{C}_{i_N} = \mathcal{C}_t)$ *be an optimal SD-MIPP solution constrained by budget* $\widetilde{\mathcal{B}}$. *Let* $\mathcal{P}$ *be the solution returned by eMIP with exponential splits of the experimental budget, started with increased budget* $N^{\log_2 \frac{3}{2}} \widetilde{\mathcal{B}}$. *Then* $\mathrm{I}(\mathcal{P}) \geq \frac{1-1/e}{1+\log N} \mathrm{I}(\mathcal{P}_C^*)$.

*Proof of Lemma 7.* The set of paths which e*MIP* considers under exponential splits – let us call them *exponential paths* – is in general a strict subset of the *linear paths* considered under linear splits. The proof of Lemma 4 indeed shows that the path returned by e*MIP* achieves at most a factor $\frac{1-1/e}{1+\log N}$ less information than the optimal exponential path. We need to show that increasing the budget by a factor of $N^{\log_2 \frac{3}{2}} \widetilde{\mathcal{B}}$ guarantees that the optimal linear path is a feasible exponential path. Every exponential path can be represented by a complete binary tree, whereby every internal node at a given level in the tree corresponds to a choice of middle node and experimental budget allocation to the left and right sub-path at the corresponding recursion level. Further, every leaf in the tree corresponds to a set of observations selected in a visited cell. Consider the tree $T^*$ representing the optimal linear path with budget $\widetilde{\mathcal{B}}$. At each inner node, the restriction to exponential splits can lead to a situation, where either the left or right sub-path receives less experimental budget than allocated by the optimal path. Our proof strategy is to turn $T^*$ into a new tree $T'$, which selects the same observations and corresponds to a valid exponential path. In order to achieve this, we will annotate each inner node $v$, which receives $B_v$ experimental budget in the optimal linear allocation, by a new feasible exponential budget $B_v' \geq B_v$. It then suffices to show that for the root $R$ it holds that $B_R' \leq (n)^{\log_2 3/2} B_R = (3/2)^{\log_2 n} B_R$. Label the edges of $T^*$ with 0 and 1, such that the sub-path corresponding to the edge labeled with 1 receives the smaller part of the linear budget split. Hence, a leaf $v$ on a path with $k$ ones receives at most $B_v \leq (1/2)^k$ of the total linear budget requirement $\widetilde{\mathcal{B}}$. Let us derive the bounds $B_v'$ bottom up. We prove by induction that $B_v' \leq (3/2)^m B_v$ where $m$ is the height of $v$ (distance from the leaves). This will suffice the condition $B_r' \leq (3/2)^{\log_2 n} B_r$, that we want to prove. For the leaves $v$ clearly $B_v' = B_v$ is sufficient, since no further split is done and hence the reward collected by both linear and exponential split will be same. Let $v$ be an inner node with children $l$ and $r$, where w.l.o.g., the left child $l$ is annotated by 0. By construction, $B_r \leq B_v/2$. By induction hypothesis, $B_l' \leq (3/2)^{m-1} B_l$, and $B_r' \leq (3/2)^{m-1} B_r$. If we choose $B_v' = B_l' + 2B_r'$, then we can find a feasible exponential budget split allocating at least $B_l'$ to $l$ and $B_r'$ to $r$. This split will require increasing the budget exponentially till we suffice $r$ and allocating the rest to $l$. To ensure that we always have a budget split that suffice $r$ with exponential budget irrespective of whether it represents $\mathcal{P}_1$ or $\mathcal{P}_2$, we need to do exponential splits from both sides, trying both exponential increase from 0 ($B_{exp}$) and $B_v - B_{exp}$ for the cases when $r$ represents $\mathcal{P}_1$ and $\mathcal{P}_2$ respectively. Now we have $B_v' \leq (3/2)^{m-1} B_l + 2(3/2)^{m-1} B_r = (3/2)^{m-1} B_v + (3/2)^{m-1} B_r \leq (3/2)^m B_v$. $\qquad\square$

**Theorem 5.** *Let* $\mathcal{P}^*$ *be the optimal solution for the single robot instance of the MIPP problem with budget constraint $B$. Then, our eSIP algorithm will find a solution* $\widehat{\mathcal{P}}$ *achieving an information value of at least* $\mathrm{I}(\widehat{\mathcal{P}}) \geq \frac{1-1/e}{1+\log_2 N} \mathrm{I}(\mathcal{P}^*)$, *whose cost is no more than* $2(2\sqrt{2}B + 4L)(1 + L\frac{\sqrt{2}}{C_{exp}})$ *in the case of linear budget split for* $\widetilde{\mathcal{B}}_e$ *and no more than* $2(2\sqrt{2}B + 4L)(1 + L\frac{\sqrt{2}}{C_{exp}})N^{\log_2 \frac{3}{2}}$ *in the case of exponential budget split for* $\widetilde{\mathcal{B}}_e$.

*Proof of Theorem 5.* Let $\tilde{B}$ be the budget requirement for SD-MIPP according to Lemma 4 (or Lemma 7 in the case of exponential splits) and $\mathcal{P}$ be the corresponding solution returned by *eMIP*. Let $C_{exp}$ be the cost of making an observation at each sensing location. Maximum number of sensing locations visited by $\mathcal{P}$ will be $\frac{\tilde{B}}{C_{exp}}$. Since we do not account for traveling to the sensing locations, an additional cost equivalent to traveling from the centroid of the visited cells to the corresponding sensing location is to be paid when the solution from SD-MIPP is transformed back to get the solution for MIPP. For each sensing location, a maximum additional cost of $L\sqrt{2}$ is incurred for traveling to the sensing location and returning back to the centroid, where $L$ is the length of the cell. Thus the additional cost for the solution path for MIPP problem, transformed from SD-MIPP problem is upper bounded by $\frac{\tilde{B}L\sqrt{2}}{C_{exp}}$. Since e*MIP* only considers exponential budget splits into traveling and experimental budget, an increase of the budget by another factor of 2 guarantees that the split defined by the optimal MIPP solution is feasible. Combining this analysis with Lemma 3 and Lemma 4 completes the proof. □

## References

Arkin, E. M., Mitchell, J. S. B., & Narasimhan, G. (1998). Resource-constrained geometric network optimization. In *Symposium on Computational Geometry*, pp. 307–316.

Bai, X., Kumar, S., Xua, D., Yun, Z., & Lai, T. H. (2006). Deploying wireless sensors to achieve both coverage and connectivity. In *Proceedings of the 7th ACM international symposium on Mobile ad hoc networking and computing*, pp. 131–142.

Batalin, M. A., Rahimi, M., Yu, Y., Liu, D., Kansal, A., Sukhatme, G. S., Kaiser, W. J., Hansen, M., Pottie, G. J., Srivastava, M., & Estrin, D. (2004). Call and response: experiments in sampling the environment. In *Proceedings of the 2nd international conference on Embedded networked sensor systems*, pp. 25–38.

Blum, A., Chawla, S., Karger, D. R., Lane, T., Meyerson, A., & Minkoff, M. (2003). Approximation algorithms for orienteering and discounted-reward tsp. In *Annual Symposium on Foundation of Computer Science (FOCS)*, p. 46.

Blum, A. L., & Furst, M. L. (1997). Fast planning through planning graph analysis. *Artificial Intelligence*, *90*, 1636–1642.

Borgstrom, P. H., Stealey, M. J., Batalin, M. A., & Kaiser, W. J. (2006). NIMS3D: A novel rapidly deployable robot for 3-dimensional applications. In *IEEE/RSJ International Conference on Intelligent Robots and Systems*, Beijing, China.

Bourgault, F., Makarenko, A., Williams, S., Grocholsky, B., & Durrant-Whyte, H. (2002). Information based adaptive robotic exploration. In *IEEE/RSJ International Conference on Intelligent Robots and Systems (IROS)*, pp. 540–545.

Brekke, L. D., Miller, N. L., Bashford, K. E., Quinn, N. W., & Dracup, J. A. (2004). Climate change impacts uncertainty for water resources in the san joaquin river basin, california. *Journal of the American water resource association*, *40*, 149–164.

Briel, M. V. D., Sanchez, R., Do, M. B., & Kambhampati, S. (2004). Effective approaches for partial satisfaction (over-subscription) planning. In *In AAAI*, pp. 562–569. AAAI Press.

Butt, S. E., & Ryan, D. M. (1999). An optimal solution procedure for the multiple tour maximum collection problem using column generation. *Computers and Operations Research*, *26*, 427–441.

Calinescu, G., Chekuri, C., Pl, M., & Vondrk, J. (2007). Maximizing a submodular set function subject to a matroid constraint (extended abstract). In *Integer Programming and Combinatorial Optimization (IPCO)*, Vol. 4513 of *Lecture Notes in Computer Science*, pp. 182–196.

Caselton, W., & Zidek, J. (1984). Optimal monitoring network design. *Statistics and Probability Letters*.

Chao, I.-M., Golden, B. L., & Wasil, E. A. (1996). A fast and effective heuristic for the orienteering problem. *European Journal of Operations Research*, *88*, 475–489.

Chekuri, C., Korula, N., & Pál, M. (2008). Improved algorithms for orienteering and related problems. In *Proc. 19th Annual ACM-SIAM Symposium on Discrete Algorithms (SODA'08)*. SIAM. To appear.

Chekuri, C., & Pal, M. (2005). A recursive greedy algorithm for walks in directed graphs. In *Annual Symposium on Foundation of Computer Science (FOCS)*, pp. 245–253.

Christofides, N. (1976). Worst-case analysis of a new heuristic for the traveling salesman problem. *Tech report,CMU*.

Cressie, N. A. C. (1991). *Statistics for Spatial Data*. Wiley.

Dhariwal, A., Zhang, B., Stauffer, B., Oberg, C., Sukhatme, G. S., Caron, D. A., & Requicha, A. A. (2006). Networked aquatic microbial observing system. In *IEEE International Conference on Robotics and Automation (ICRA)*.

Feillet, D., Dejax, P., & Gendreau, M. (2005). Traveling salesman problem with profits. *Transportation Science*, *39*(2), 188–205.

Garg, N. (2005). Saving an epsilon: a 2-approximation for the k-mst problem in graphs. In *ACM Symposium on Theory of Computing (STOC)*, pp. 396–402.

Golden, B., Levy, L., & Vohra, R. (1987). The orienteering problem. *Naval Research Logistics*, *34*, 307–318.

Goundan, P. R., & Schulz, A. S. (2008). Revisiting the greedy approach to submodular set function maximization.. Working paper, MIT.

Guestrin, C., Krause, A., & Singh, A. P. (2005). Near-optimal sensor placements in gaussian processes. In *International Conference on Machine Learning (ICML)*.

Harmon, T. C., Ambrose, R. F., Gilbert, R. M., Fisher, J. C., Stealey, M., & Kaiser, W. J. (2007). High-resolution river hydraulic and water quality characterization using rapidly deployable networked infomechanical systems (NIMS RD). *Environmental Engineering Science*, *24*(2), 151–159.

I-Ming, C., Golden, B., & Wasil, E. (1996). The team orienteering problem. *European Journal of Operation Research*, *88*, 464–474.

Ibaraki, T., Muro, S., Murakami, T., & Hasegawa, T. (1983). Using branch-and-bound algorithms to obtain suboptimal solutions. *Mathematical Methods of Operations Research*, *27*(1), 177–202.

Ishikawa, T., & Tanaka, M. (1993). Diurnal stratification and its effects on wind-induced currents and water qualities in lake kasumigaura, japan. *Journal of Hydraulic Research*, *31*(3), 307–322.

Johnson, D. S., Minkoff, M., & Phillips, S. (2000). The prize collecting steiner tree problem: theory and practice. In *Symposium on Discrete Algorithms (SODA)*, pp. 760–769.

Jordan, B. L., Batalin, M. A., & Kaiser, W. J. (2007). NIMS RD: A rapidly deployable cable based robot. In *IEEE International Conference on Robotics and Automation (ICRA)*, Rome, Italy.

Kataoka, S., & Morito, S. (1988). An algorithm for the single constraint maximum collection problem. *Journal of the Operational Research Society of Japan*, *31*, 515–530.

Ko, C.-W., Lee, J., & Queyranne, M. (1995). An exact algorithm for maximum entropy sampling. *Operations Research*, *43*(4), 684–691.

Krause, A., & Guestrin, C. (2007). Near-optimal observation selection using submodular functions. In *AAAI Nectar track*.

Krause, A., Singh, A., & Guestrin, C. (2008). Near-optimal sensor placements in Gaussian processes: Theory, efficient algorithms and empirical studies. In *Journal of Machine Learning and Research (JMLR)*, Vol. 9, pp. 235–284.

Krause, A., & Guestrin, C. (2007). Nonmyopic active learning of gaussian processes: an exploration-exploitation approach. In *International Conference on Machine Learning (ICML)*, pp. 449–456.

Krause, A., Guestrin, C., Gupta, A., & Kleinberg, J. (2006). Near-optimal sensor placements: Maximizing information while minimizing communication cost. In *Proceedings of the fifth international conference on Information processing in sensor networks (IPSN)*, pp. 2–10.

Laporte, G., & Martello, S. (1990). The selective travelling salesman problem. *Discrete Applied Mathematics*, *26*, 193–207.

Lau, H., Huang, S., & Dissanayake, G. (2006). Probabilistic search for a moving target in an indoor environment. In *IEEE/RSJ International Conference on Intelligent Robots and Systems (IROS)*, pp. 3393–3398.

Lin, S. (1965). Computer solutions of the traveling salesman problem. *Bell System Technical Journal*, *44*, 2245–2269.

MacIntyre, S. (1993). Vertical mixing in a shallow, eutrophic lake: Possible consequences for the light climate of phytoplankton. *Limnology and Oceanography*, *38*(4), 798–817.

MacIntyre, S., Romero, J. R., & Kling, G. W. (2002). Spatial-temporal variability in surface layer deepening and lateral advection in an embayment of lake victoria, east africa. *Limnology and Oceanography*, *47*(3), 656–671.

Meliou, A., Krause, A., Guestrin, C., & Hellerstein, J. M. (2007). Nonmyopic informative path planning in spatio-temporal models. In *Association for Advancement of Artificial Intelligence (AAAI)*, pp. 602–607.

Nagarajan, V., & Ravi, R. (2007). Poly-logarithmic approximation algorithms for directed vehicle routing problems. In *Proc. 10th Internat. Workshop on Approximation Algorithms for Combinatorial Optimization Problems (APPROX'07)*, Vol. 4627 of *LNCS*, pp. 257–270. Springer.

Nemhauser, G., Wolsey, L., & Fisher, M. (1978). An analysis of the approximations for maximizing submodular set functions. *Mathematical Programming*, *14*, 265–294.

Pon, R., Batalin, M., Gordon, J., Rahimi, M., Kaiser, W., Sukhatme, G., Srivastava, M., & Estrin, D. (2005). Networked infomechanical systems: A mobile wireless sensor network platform. In *Proceedings of the fifth international conference on Information processing in sensor networks (IPSN)*, pp. 376–381.

Rahimi, M., Pon, R., Kaiser, W., Sukhatme, G., Estrin, D., & Srivastava, M. (2004). Adaptive sampling for environmental robotics. In *IEEE International Conference on Robotics and Automation (ICRA)*.

Rasmussen, C. E., & Williams, C. K. (2006). *Gaussian Process for Machine Learning*. Adaptive Computation and Machine Learning. MIT Press.

Reynolds-Fleming, J. V., Fleming, J. G., & Luettich, R. A. (2004). Portable autonomous vertical profiler for estuarine applications. *Estuaries*, *25*, 142–147.

Roy, N., & Earnest, C. (2006). Dynamic action spaces for information gain maximization in search and exploration. In *American Control Conference*.

Ryan, M. R. K. (2008). Exploiting subgraph structure in multi-robot path planning. In *Journal of Artificial Intelligence and Research (JAIR)*, Vol. 31, pp. 497–542.

Sim, R., & Roy, N. (2005). Global a-optimal robot exploration in slam. In *IEEE International Conference on Robotics and Automation (ICRA)*.

Simmons, R. G., Apfelbaum, D., Burgard, W., Fox, D., Moors, M., Thrun, S., & Younes, H. (2000). Coordination for multi-robot exploration and mapping. In *Association for Advancement of Artificial Intelligence (AAAI)*, pp. 852–858.

Singh, A., Nowak, R., & Ramanathan, P. (2006). Active learning for adaptive mobile sensing networks. In *Proceedings of the fifth international conference on Information processing in sensor networks (IPSN)*, pp. 60–68.

Singh, A., Batalin, M. A., Chen, V., Stealey, M. J., Jordan, B., Fisher, J., Harmon, T., Hansen, M., & Kaiser, W. J. (2007a). Autonomous robotic sensing experiments at san joaquin river. In *IEEE International Conference on Robotics and Automation (ICRA)*, pp. 4987–4993, Rome, Italy.

Singh, A., Krause, A., Guestrin, C., Kaiser, W. J., & Batalin, M. A. (2007b). Efficient planning of informative paths for multiple robots. In *International Joint Conference on Artificial Intelligence (IJCAI)*, pp. 2204–2211, Hyderabad, India.

Smith, D. E. (2004). Choosing objectives in over-subscription planning. In *International Conference on Automated Planning and Scheduling (ICAPS)*.

Somasundara, A. A., Ramamoorthy, A., & Srivastava, M. B. (2007). Mobile element scheduling with dynamic deadlines. In *IEEE Transactions on Mobile Computing*, Vol. 6, pp. 395–410.

Stachniss, C., Grisetti, G., & Burgard, W. (2005). Information gain-based exploration using rao-blackwellized particle filters. In *Robotics Science and Systems (RSS)*.

Tang, L., & Wang, X. (2006). Iterated local search algorithm based on very large-scale neighborhood for prize-collecting vehicle routing problem. *The International Journal of Advanced Manufacturing Technology*, 1–13.

Thompson, D. R., & Wettergreen, D. (2008). Intelligent maps for autonomous kilometer-scale science survey. In *International Symposium on Artificial Intelligence, Robotics and Automation in Space (iSAIRAS)*.

Trummel, K. E., & Weisinger, J. R. (1986). The complexity of the optimal searcher path problem. *Operations Research*, *34*(2), 324–327.

Zhang, W., & Korf, R. E. (1995). Performance of linear-space search algorithms. *Artificial Intelligence*, *79*(2), 241–292.